\documentclass[12pt]{article}
 
\usepackage{amsmath}
\usepackage{times}
\usepackage{graphicx}
\usepackage{color}
\usepackage{multirow}
\usepackage[authoryear]{natbib}
\usepackage{rotating}
\usepackage{bbm}
\usepackage{latexsym}

\usepackage{algorithm}
\usepackage{algorithmic}
\usepackage{subfigure} 
\usepackage{amsfonts}
\usepackage{amssymb}
\usepackage{amsthm}
\usepackage{url}

\newtheorem{theorem}{Theorem}
\newtheorem*{theorem*}{Theorem}

\newcommand\independent{\protect\mathpalette{\protect\independenT}{\perp}}
\def\independenT#1#2{\mathrel{\rlap{$#1#2$}\mkern2mu{#1#2}}}

\newcommand{\argmin}{\mathop{\mathrm{argmin}}}

\newcommand{\bx}{\boldsymbol{x}}
\newcommand{\by}{\boldsymbol{y}}
\newcommand{\bz}{\boldsymbol{z}}
\newcommand{\bX}{\boldsymbol{X}}
\newcommand{\bY}{\boldsymbol{Y}}
\newcommand{\bZ}{\boldsymbol{Z}}
\newcommand{\bI}{\boldsymbol{I}}
\newcommand{\bzp}{\boldsymbol{z}_{\perp}}
\newcommand{\dx}{\mathrm{d}\boldsymbol{x}}
\newcommand{\dy}{\mathrm{d}\boldsymbol{y}}
\newcommand{\dz}{\mathrm{d}\boldsymbol{z}}
\newcommand{\dzp}{\mathrm{d}\boldsymbol{z}_{\perp}}
\newcommand{\bv}{\boldsymbol{v}}
\newcommand{\bu}{\boldsymbol{u}}
\newcommand{\but}{\boldsymbol{\tilde{u}}}
\newcommand{\bW}{\boldsymbol{W}}

\newcommand{\bWp}{\boldsymbol{W}_{\perp}}
\newcommand{\bhW}{\widehat{\boldsymbol{W}}}

\newcommand{\bs}{\boldsymbol{s}}
\newcommand{\ba}{\boldsymbol{a}}

\newcommand{\talpha}{\widetilde{\alpha}}
\newcommand{\balpha}{\boldsymbol{\alpha}}
\newcommand{\bhalpha}{\widehat{\balpha}}
\newcommand{\btalpha}{\widetilde{\balpha}}
\newcommand{\bhbeta}{\widehat{\boldsymbol{\beta}}}

\newcommand{\bvphi}{\boldsymbol{\varphi}}
\newcommand{\bphib}{\bar{\boldsymbol{\Phi}}}

\newcommand{\bhG}{\widehat{\boldsymbol{G}}}
\newcommand{\bhh}{\widehat{\boldsymbol{h}}}
\newcommand{\hQ}{\widehat{G}}
\newcommand{\hq}{\widehat{h}}

\newcommand{\px}{p(\bx)}
\newcommand{\py}{p(\by)}
\newcommand{\pz}{p(\bz)}
\newcommand{\pxy}{p(\bx,\by)}
\newcommand{\pzy}{p(\bz,\by)}
\newcommand{\pzzp}{p(\bz,\bzp)}
\newcommand{\pzzpy}{p(\bz,\bzp,\by)}
\newcommand{\pygx}{p(\by|\bx)}
\newcommand{\pygz}{p(\by|\bz)}

\newcommand{\pzpgz}{p(\bzp|\bz)}

\newcommand{\hSCE}{\widehat{\mathrm{SCE}}}
\newcommand{\pSCEW}{\frac{\partial \hSCE}{\partial W_{l,l'}}}
\newcommand{\pSCEbW}{\frac{\partial \hSCE}{\partial \bW}}
\newcommand{\pa}{\partial}
\newcommand{\paW}{\partial W_{l,l'}}

\title{
Conditional Density Estimation \\with Dimensionality Reduction 
via \\Squared-Loss Conditional Entropy Minimization
}

\author{
Voot Tangkaratt, Ning Xie, and Masashi Sugiyama\\[2mm]
Tokyo Institute of Technology, Japan.\\
\{voot@sg., xie@sg., sugi@\}cs.titech.ac.jp
}

\date{}

\begin{document} 
\maketitle

\begin{abstract}
Regression aims at estimating the conditional mean of output given input.
However, regression is not informative enough if the conditional density
is multimodal, heteroscedastic, and asymmetric.
In such a case, estimating the conditional density itself is preferable,
but conditional density estimation (CDE) is challenging in high-dimensional space.
A naive approach to coping with high-dimensionality
is to first perform dimensionality reduction (DR) and then execute CDE.
However, such a two-step process does not perform well in practice
because the error incurred in the first DR step can be magnified in the second CDE step.
In this paper, we propose a novel single-shot procedure 
that performs CDE and DR simultaneously in an integrated way.
Our key idea is to formulate DR
as the problem of minimizing a squared-loss variant of conditional entropy,
and this is solved via CDE.
Thus, an additional CDE step is not needed after DR.
We demonstrate the usefulness of the proposed method
through extensive experiments on various datasets including
humanoid robot transition and computer art.
\end{abstract} 


\section{Introduction} \label{chap:intro}
Analyzing input-output relationship from samples
is one of the central challenges in machine learning.
The most common approach is \emph{regression}, which estimates
the conditional mean of output $\by$ given input $\bx$.
However, just analyzing the conditional mean is not informative enough,
when the conditional density $\pygx$ possesses
multimodality, asymmetry, and heteroscedasticity (i.e., input-dependent variance)
as a function of output $\by$.
In such cases, it would be more appropriate to
estimate the conditional density itself
(Figure~\ref{fig:illustration}).

The most naive approach to conditional density estimation (CDE) would be 
\emph{$\epsilon$-neighbor kernel density estimation} ($\epsilon$-KDE),
which performs standard KDE along $\by$ only with nearby samples in the input domain.
However, $\epsilon$-KDE do not work well in high-dimensional problems
because the number of nearby samples is too few.
To avoid the small sample problem, KDE may be applied twice
to estimate $\pxy$ and $\px$ separately and the estimated densities
may be plugged into the decomposed form $\pygx=\pxy/\px$
to estimate the conditional density.
However, taking the ratio of two estimated densities significantly magnifies
the estimation error and thus is not reliable.
To overcome this problem, an approach
to directly estimating the density ratio $\pxy/\px$
without separate estimation of densities $\pxy$ and $\px$ has been explored
\citep{AISTATS:Sugiyama+etal:2010}.
This method, called \emph{least-squares CDE} (LSCDE),
was proved to possess the optimal non-parametric learning rate in the mini-max sense,
and its solution can be efficiently and analytically computed.
Nevertheless, estimating conditional densities in high-dimensional problems
is still challenging.

A natural idea to cope with the high-dimensionality
is to perform \emph{dimensionality reduction} (DR) before CDE.
\emph{Sufficient DR}
\citep{JASA:Li:1991,JASA:Cook+Ni:2005} is a framework of supervised DR
aimed at finding the subspace of input $\bx$ that contains all information on output $\by$,
and a method based on conditional-covariance operators in reproducing kernel Hilbert spaces
has been proposed \citep{AS:Fukumizu+etal:2009}.
Although this method possesses superior thoretical properties,
it is not easy to use in practice
because no systematic model selection method
is available for kernel parameters.
To overcome this problem, an alternative sufficient DR method based on 
\emph{squared-loss mutual information} (SMI)
has been proposed recently \citep{NC:Suzuki+Sugiyama:2013}.
This method involves non-parametric estimation of SMI 
that is theoretically guaranteed to achieve the optimal estimation rate,
and all tuning parameters can be systematically chosen in practice
by cross-validation with respect to the SMI approximation error.

Given such state-of-the-art DR methods, 
performing DR before LSCDE would be a promising approach
to improving the accuracy of CDE in high-dimensional problems.
However, such a two-step approach is not preferable
because DR in the first step is performed without regard to CDE in the second step
and thus small error incurred in the DR step can be significantly magnified
in the CDE step.

\begin{figure}[t]
  \centering	
  \subfigure[CDE without DR]{
  \includegraphics[width=0.80\textwidth]{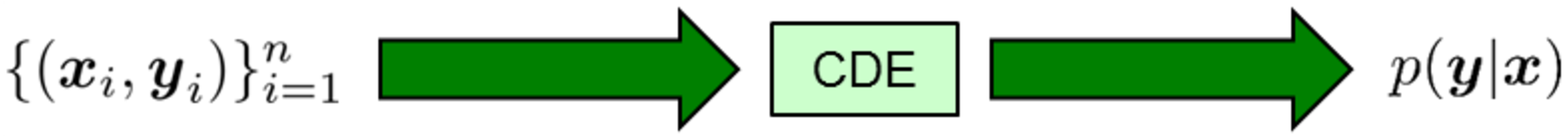}
  }
  \subfigure[CDE after DR]{
  \includegraphics[width=0.80\textwidth]{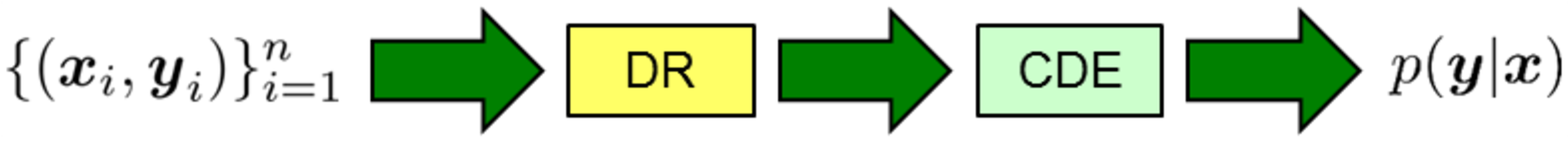}
  }
  \subfigure[CDE with DR (proposed)]{
  \includegraphics[width=0.80\textwidth]{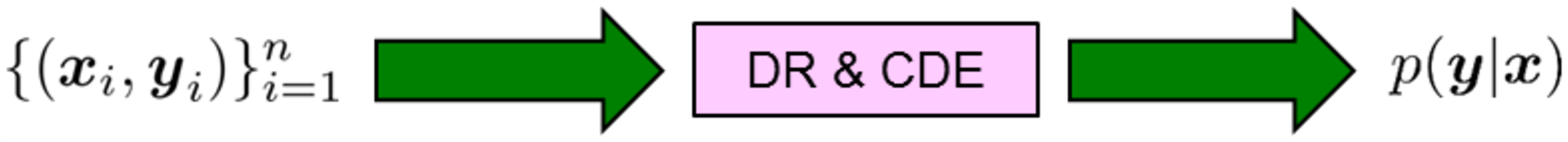}
  }
  \caption{Conditional density estimation (CDE) and dimensionality reduction (DR).
    (a) CDE without DR performs poorly in high-dimensional problems.
    (b) CDE after DR can magnify the small DR error in the CDE step.
    (c) CDE with DR (proposed) performs CDE in the DR process in an integrated manner.}
	\label{fig:illust_problem}
\end{figure}

In this paper, we propose a single-shot method that integrates DR and CDE.
Our key idea is to formulate the sufficient DR problem in terms of 
the \emph{squared-loss conditional entropy} (SCE)
which includes the conditional density in its definition,
and LSCDE is executed when DR is performed.
Therefore, when DR is completed, the final conditional density estimator
has already been obtained without an additional CDE step (Figure~\ref{fig:illust_problem}).
We demonstrate the usefulness of the proposed method,
named \emph{least-squares conditional entropy} (LSCE),
through experiments on benchmark datasets, humanoid robot control simulations, and computer art.

\section{Conditional Density Estimation with Dimensionality Reduction} \label{chap:proposed}
In this section, we describe our proposed method
for conditional density estimation with dimensionality reduction.

\subsection{Problem Formulation}
Let $\mathcal{D}_{\textbf{x}}(\subset \mathbb{R}^{d_\textbf{x}})$ and
$\mathcal{D}_{\textbf{y}}(\subset \mathbb{R}^{d_\textbf{y}})$ be
the input and output domains
with dimensionality $d_\textbf{x}$ and $d_\textbf{y}$, respectively,
and let $\pxy$ be a joint probability density on $\mathcal{D}_{\textbf{x}} \times \mathcal{D}_{\textbf{y}}$.
Assume that we are given $n$ independent and identically distributed (i.i.d.)
training samples from the joint density: 
\[
\{ (\bx_i, \by_i) \}_{i=1}^n \stackrel{\mathrm{i.i.d.}}{\sim} \pxy.
\] 
The goal is to estimate the conditional density $\pygx$ from the samples.

Our implicit assumption is that the input dimensionality $d_\textbf{x}$ is large,
but its ``intrinsic'' dimensionality, denoted by $d_\textbf{z}$, is rather small.
More specifically, let $\bW$ and $\bWp$ be $d_\textbf{z}\times d_\textbf{x}$
and $(d_\textbf{x}-d_\textbf{z})\times d_\textbf{x}$ matrices
such that $\begin{bmatrix}\bW^\top, \bWp^\top\end{bmatrix}$ is an orthogonal matrix.
Then we assume that $\bx$ can be decomposed into
the component $\bz=\bW\bx$ and its perpendicular component $\bzp = \bWp\bx$ so that
$\by$ and $\bx$ are conditionally independent given $\bz$:
\begin{align}
\by \independent \bx | \bz.
\label{by-ind-bx|bz}
\end{align}
This measn that $\bz$ is the relevant part of $\bx$,
and the rest $\bzp$ does not contain any information on $\by$.
The problem of finding $\bW$ is called \emph{sufficient dimensionality reduction}
\citep{JASA:Li:1991,JASA:Cook+Ni:2005}.

\subsection{Sufficient Dimensionality Reduction with SCE}
Let us consider a squared-loss variant of 
conditional entropy named \emph{squared-loss CE} (SCE):
\begin{align}
\mathrm{SCE}(\bY|\bZ) &= -\frac{1}{2} \iint \Big( \pygz - 1 \Big)^2 \pz \dz \dy. \label{SCE:eq}
\end{align}
By expanding the squared term in Eq.\eqref{SCE:eq}, we obtained
\begin{align}
\mathrm{SCE}(\bY|\bZ) &= -\frac{1}{2} \iint \pygz^2\pz\dz\dy + \iint \pygz \pz\dz\dy - \frac{1}{2}\iint \pz \dz \dy \nonumber \\
  &= -\frac{1}{2} \iint \pygz^2\pz\dz\dy + 1 - \frac{1}{2}\int \dy \nonumber \\
  &= \widetilde{\mathrm{SCE}}(\bY|\bZ) + 1 - \frac{1}{2}\int \dy,
\end{align}
where $\widetilde{\mathrm{SCE}}(\bY|\bZ)$ is defined as 
\begin{align}
\widetilde{\mathrm{SCE}}(\bY|\bZ) &= -\frac{1}{2} \iint \pygz^2\pz\dz\dy. \label{SCE:eqt}
\end{align}
Then we have the following theorem
(its proof is given in Appendix~\ref{proof:SCE(Y|Z)-SCE(Y|X)>=0}),
which forms the basis of our proposed method:
\begin{theorem} \label{theorem:SCE(Y|Z)-SCE(Y|X)>=0}
\begin{align*}
\widetilde{\mathrm{SCE}}(\bY|\bZ) - \widetilde{\mathrm{SCE}}(\bY|\bX)
 &=  \frac{1}{2}\iint \left( \frac{p(\bzp,\by|\bz) }{\pzpgz \pygz } - 1 \right)^2
 \pygz^2 \px \dx \dy \\
 &\geq 0.
\end{align*} 
\end{theorem}

This theorem shows $\widetilde{\mathrm{SCE}}(\bY|\bZ) \ge \widetilde{\mathrm{SCE}}(\bY|\bX)$,
and the equality holds if and only if 
\begin{align*}
p(\bzp,\by|\bz) = \pzpgz \pygz.
\end{align*}
This is equivalent to the conditional independence \eqref{by-ind-bx|bz},
and therefore sufficient dimensionality reduction can be performed
by minimizing $\widetilde{\mathrm{SCE}}(\bY|\bZ)$ with respect to $\bW$:
\begin{align}
  \bW^*=\argmin_{\bW\in\mathbb{G}_{d_\textbf{z}}^{d_\textbf{x}}(\mathbb{R})}\widetilde{\mathrm{SCE}}(\bY|\bZ=\bW\bX).
\label{bW*}
\end{align}
Here, $\mathbb{G}_{d_\textbf{z}}^{d_\textbf{x}}(\mathbb{R})$ denotes the \emph{Grassmann manifold},
which is a set of orthogonal matrices without overlaps:
\begin{align*}
  \mathbb{G}_{d_\textbf{z}}^{d_\textbf{x}}(\mathbb{R})
  = \{\bW \in \mathbb{R}^{d_\textbf{z} \times d_\textbf{x}} \ | \ \bW\bW^\top = \bI_{d_\textbf{z}}\}/\sim,
\end{align*}
where $\bI$ denotes the identity matrix and
$\sim$ represents the equivalence relation:
$\bW$ and $\bW'$ are written as $\bW \sim \bW'$ if their rows span the same subspace.

Since $\pygz=\pzy/\pz$,
$\mathrm{SCE}(\bY|\bZ)$ is equivalent to the negative \emph{Pearson divergence}
\citep{PhMag:Pearson:1900} from $\pzy$ to $\pz$,
which is a member of the \emph{$f$-divergence} class \citep{JRSS-B:Ali+Silvey:1966,SSM-Hungary:Csiszar:1967}
with the squared-loss function.
On the other hand, ordinary conditional entropy (CE), defined by
\begin{align*}
\mathrm{CE}(\bY|\bZ) = -\iint \pzy \log \pygz \dz \dy,
\end{align*}
is the negative \emph{Kullback-Leibler divergence}
\citep{Annals-Math-Stat:Kullback+Leibler:1951} from $\pzy$ to $\pz$.
Since the Kullback-Leibler divergence is also a member of the $f$-divergence class
(with the log-loss function), CE and SCE have similar properties.
Indeed, the above theorem also holds for ordinary CE.
However, the Pearson divergence is shown to be more robust
against outliers \citep{AISM:Sugiyama+etal:2012},
since the log function---which is very sharp near zero---is not included.
Furthermore, as shown below, $\widetilde{\mathrm{SCE}}$ can be approximated \emph{analytically}
and thus its derivative can also be easily computed.
This is a critical property for developing a dimensionality reduction method
because we want to minimize $\widetilde{\mathrm{SCE}}$ with respect to $\bW$,
where the gradient is highly useful in devising an optimization algorithm.
For this reason, we adopt SCE instead of CE below.

\subsection{SCE Approximation}
\label{subsection:SCE_LSCDE}
Since $\widetilde{\mathrm{SCE}}(\bY|\bZ)$ in Eq.\eqref{bW*} is unknown in practice,
we approximate it using samples $\{ (\bz_i, \by_i) ~|~\bz_i=\bW\bx_i\}_{i=1}^n$.

The trivial inequality $(a-b)^2/2\ge0$ yields $a^2/2\ge ab - b^2/2$, and thus we have
\begin{align}
  \frac{a^2}{2}=\max_b \left[ab - \frac{b^2}{2}\right].
  \label{a^2/2=max_b[ab-b^2/2]}
\end{align}
If we set $a=\pygz$, we have
\begin{align*}
   \frac{\pygz ^2}{2}
  \ge \max_b \left[\pygz b(\bz,\by)-\frac{b(\bz,\by)^2}{2}\right].
\end{align*}
If we multiply both sides of the above inequality
with $-\pz$, and integrated over $\bz$ and $\by$, we have
\begin{align}
\widetilde{\mathrm{SCE}}(\bY|\bZ) \le \min_b \iint\left[ \frac{b(\bz,\by)^2\pz}{2}-b(\bz,\by)\pzy\right] \dz \dy,
\label{SCE-upperbound}
\end{align}
where minimization with respect to $b$ is now performed as a function of $\bz$ and $\by$.
For more general discussions on divergence bounding,
see \citep{CRM:Keziou:2003} and \citep{IEEE-IT:Nguyen+etal:2010}.

Let us consider a linear-in-parameter model for $b$:
\begin{align*}
  b(\bz,\by)=\balpha^\top\bvphi(\bz, \by),
\end{align*}
where $\balpha$ is a parameter vector and $\bvphi(\bz, \by)$ is a vector of basis functions.
If the expectations over densities $\pz$ and $\pzy$ are approximated by samples averages
and the $\ell_2$-regularizer $\lambda\balpha^\top\balpha/2$ ($\lambda\ge0$) is included,
the above minimization problem yields
\begin{align*}
  \bhalpha=\argmin_{\balpha}   
  \left[\frac{1}{2}\balpha^\top\bhG\balpha-\bhh^\top\balpha
    +\frac{\lambda}{2}\balpha^\top\balpha\right],
\end{align*}
where
\begin{align}
&\bhG = \frac{1}{n} \sum_{i=1}^n \bphib(\bz_i), \notag \\
&\bhh = \frac{1}{n} \sum_{i=1}^n \bvphi(\bz_i, \by_i),  \notag \\
&\bphib(\bz) = \int \bvphi(\bz, \by) \bvphi(\bz, \by)^\top \dy.
\label{SCE:matrices}
\end{align}
The solution $\bhalpha$ is analytically given by
\begin{align*}
  \bhalpha &= \left( \bhG + \lambda \bI \right)^{-1} \bhh,
\end{align*}
which yields $\widehat{b}(\bz,\by)=\bhalpha^\top\bvphi(\bz, \by)$.
Then, from Eq.\eqref{SCE-upperbound},
an approximator of $\widetilde{\mathrm{SCE}}(\bY|\bZ)$ is obtained analytically as
\begin{align*}
  \hSCE(\bY|\bZ)=\frac{1}{2}\bhalpha^\top\bhG\bhalpha-\bhh^\top\bhalpha.
\end{align*}
We call this method \emph{least-squares conditional entropy} (LSCE).

\subsection{Model Selection by Cross-Validation}
\label{section:model}
The above $\widetilde{\mathrm{SCE}}$ approximator depends on the choice of models,
i.e., the basis function $\bvphi(\bz, \by)$ and the regularization parameter $\lambda$.
Such a model can be objectively selected by cross-validation as follows:
\begin{enumerate}
\item 
The training dataset $\mathcal{S}=\{ (\bx_i, \by_i) \}_{i=1}^n$
is divided into $K$ disjoint subsets $\{\mathcal{S}_j\}_{j=1}^K$
with (approximately) the same size.
\item \textbf{For} each model $M$ in the candidate set,
\begin{enumerate}
\item
\textbf{For} $j=1,\ldots,K$,
\begin{enumerate}
\item 
For model $M$, the LSCE solution $\widehat{b}^{(M,j)}$ is computed from
$\mathcal{S} \backslash \mathcal{S}_j$ (i.e., all samples except $\mathcal{S}_j$).
\item 
Evaluate the upper bound of $\widetilde{\mathrm{SCE}}$ obtained by $\widehat{b}^{(M,j)}$ 
using the hold-out data $\mathcal{S}_j$:
\begin{align*}
 \mathrm{CV}_j (M) &= \frac{1}{2 |\mathcal{S}_j|}
\sum_{\bz \in \mathcal{S}_j} \int\widehat{b}^{(M,j)}(\bz,\by)^2\dy - \frac{1}{|\mathcal{S}_j|}\sum_{ (\bz, \by) \in \mathcal{S}_j} 
\widehat{b}^{(M,j)}(\bz,\by),
\end{align*}
where $|\mathcal{S}_j|$ denotes the cardinality of $\mathcal{S}_j$.
\end{enumerate}
\item The average score is computed as
\begin{align*}
\mathrm{CV}(M)= \frac{1}{K} \sum_{j=1}^K \mathrm{CV}_j(M).
\end{align*}
\end{enumerate}
\item The model that minimizes the average score is chosen:
  \begin{align*}
    \widehat{M}=\argmin_M \mathrm{CV}(M).
  \end{align*}
\item For the chosen model $\widehat{M}$,
  the LSCE solution $\widehat{b}$ is computed from all samples $\mathcal{S}$
  and the approximator $\hSCE(\bY|\bZ)$ is computed.
\end{enumerate}
In the experiments, we use $K=5$.

\subsection{Dimensionality Reduction with SCE}
Now we solve the following optimization problem by gradient descent:
\begin{align}
\argmin_{\bW \in \mathbb{G}_{d_\textbf{z}}^{d_\textbf{x}}(\mathbb{R})} \hSCE(\bY|\bZ=\bW\bX).
\end{align}
As shown in Appendix~\ref{appendix:derivative-SCE},
the gradient of $\hSCE(\bY|\bZ=\bW\bX)$ is given by
\begin{align*}
  \pSCEW&= \bhalpha^\top \frac{\pa \bhG}{\paW}\left(\frac{3}{2}\bhalpha - \bhbeta\right)
	+ \frac{\pa \bhh^\top}{\paW} (\bhbeta - 2\bhalpha), 
\end{align*}
where $\bhbeta = \left(\bhG + \lambda \bI \right)^{-1}\bhG \bhalpha$.

In the Euclidean space, the above gradient gives the steepest direction.
However, on a manifold, the \emph{natural gradient} \citep{nc:Amari:1998}
gives the steepest direction.

The natural gradient $\nabla\hSCE(\bW)$ at $\bW$
is the projection of the ordinary gradient $\pSCEW$ to the tangent space of
$\mathbb{G}_{d_\textbf{z}}^{d_\textbf{x}}(\mathbb{R})$ at $\bW$.
If the tangent space is equipped with the canonical metric
$\left\langle\bW, \bW'\right\rangle = \frac{1}{2}\mathrm{tr}(\bW^\top\bW')$,
the natural gradient is given as follows \citep{SIAM-MAA:Edelman+etal:1998}:
\begin{align*}
\nabla\hSCE = \pSCEbW - \pSCEbW \bW^\top\bW = \pSCEbW\bWp^\top\bWp,
\end{align*}
where $\bWp$ is a $(d_\textbf{x}-d_\textbf{z})\times d_\textbf{x}$ matrix such that
$\begin{bmatrix}\bW^\top, \bWp^\top\end{bmatrix}$ is an orthogonal matrix.

Then the \emph{geodesic} from $\bW$ to the direction of the natural gradient $\nabla\hSCE$
over $\mathbb{G}_{d_\textbf{z}}^{d_\textbf{x}}(\mathbb{R})$ can be expressed
using $t\in\mathbb{R}$ as
\begin{align*}
\bW_t &= \begin{bmatrix}
\boldsymbol{I}_{d_\textbf{z}} & \boldsymbol{O}_{d_\textbf{z}, (d_\textbf{x}-d_\textbf{z})}
\end{bmatrix} 
\times
\exp\left( -t \begin{bmatrix}
\boldsymbol{O}_{d_\textbf{z}, d_\textbf{z}} & \pSCEbW \bWp^\top \\
-\bWp\pSCEbW^\top & \boldsymbol{O}_{d_\textbf{x}-d_\textbf{z}, d_\textbf{x}-d_\textbf{z}}
\end{bmatrix}
\right)
\begin{bmatrix}
\bW \\
\bWp
\end{bmatrix},
\end{align*}
where ``$\exp$'' for a matrix denotes the matrix exponential
and $\boldsymbol{O}_{d,d'}$ denotes the $d\times d'$ zero matrix.
Note that the derivative $\partial_t\bW_t$ at $t=0$ coincides with the natural gradient
$\nabla\hSCE$; see \citep{SIAM-MAA:Edelman+etal:1998} for details.
Thus, line search along the geodesic
in the natural gradient direction is equivalent to finding the minimizer from $\{\bW_t~|~t\ge0\}$.

Once $\bW$ is updated, SCE is re-estimated with the new $\bW$ and gradient descent is performed again.
This entire procedure is repeated until $\bW$ converges.
When SCE is re-estimated, performing cross-validation in every step is computationally expensive.
In our implementation, we perform cross-validation only once every 5 gradient updates.
Furthermore, to find a better local optimal solution, this gradient descent procedure is
executed 20 times with randomly chosen initial solutions
and the one achieving the smallest value of $\hSCE$ is chosen.

\subsection{Conditional Density Estimation with SCE}
Since the maximum of Eq.\eqref{a^2/2=max_b[ab-b^2/2]}
is attained at $b=a$ and $a=\pygz$ in the current derivation,
the optimal $b(\bz,\by)$ is actually the conditional density $\pygz$ itself.
Therefore, $\bhalpha^\top\bvphi(\bz, \by)$ obtained by LSCE
is a conditional density estimator.
This actually implies that the upper-bound minimization procedure 
described in Section~\ref{subsection:SCE_LSCDE} is 
equivalent to \emph{least-squares conditional density estimation} (LSCDE)
\citep{AISTATS:Sugiyama+etal:2010},
which minimizes the squared error:
\begin{align*}
  \frac{1}{2}\iint\Big(b(\bz,\by)-\pygz\Big)^2\pz\dz\dy.
\end{align*}
Then, in the same way as the original LSCDE,
we may post-process the solution $\bhalpha$
to make the conditional density estimator non-negative and normalized as
\begin{align}
\widehat{p}(\by|\bz = \widetilde{\bz}) 
= \frac{\btalpha^\top\bvphi(\widetilde{\bz}, \by)}
{\int \btalpha^\top\bvphi(\widetilde{\bz}, \by') \dy'},
\label{LSCDE:normalize}
\end{align}
where $\widetilde{\alpha}_l= \max \left( \widehat{\alpha}_l, 0 \right)$.
Note that, even if the solution is post-processed as Eq.\eqref{LSCDE:normalize},
the optimal estimation rate of the LSCDE solution is still maintained \citep{AISTATS:Sugiyama+etal:2010}.

\subsection{Basis Function Design}
In practice, we use the following Gaussian function as the $k$-th basis:
\begin{align}
\varphi_k(\bz, \by)
   &= \exp \left( - \frac{\| \bz - \bu_k \|^2+\| \by - \bv_k \|^2}{2 \sigma^2} \right) ,
\end{align}
where $(\bu_k,\bv_k)$ denotes the $k$-th Gaussian center
located at $(\bz_k, \by_k )$.
When the sample size $n$ is too large,
we may use only a subset of samples as Gaussian centers.
$\sigma$ denotes the Gaussian bandwidth, which is chosen by cross-validation
as explained in Section~\ref{section:model}. 
We may use different bandwidths for $\bz$ and $\by$,
but this will increase the computation time for model selection.
In our implementation, we normalize each element of $\bz$ and $\by$
to have the unit variance in advance and then use the common bandwidth for $\bz$ and $\by$.

A notable advantage of using the Gaussian function is that
the integral over $\by$ appeared in $\bphib(\bz)$ (see Eq.\eqref{SCE:matrices})
can be computed analytically as
\begin{align*}
\bar{\Phi}_{k, k'}(\bz)
  &= (\sqrt{\pi}\sigma)^{d_\textbf{y}} \
\exp \left( - \frac{2\| \bz - \bu_k \|^2+2\| \bz - \bu_{k'} \|^2 + \| \bv_k - \bv_{k'} \|^2}{4 \sigma^2} \right). 
\end{align*}
Similarly,  the normalization term in Eq.\eqref{LSCDE:normalize} 
can also be computed analytically as
\begin{align*}
  \int \btalpha^\top\bvphi(\bz, \by) \dy &
= (\sqrt{ 2 \pi}\sigma)^{d_\textbf{y}}
\sum_k \talpha_k 
\exp \left( - \frac{\| \bz - \bu_k \|^2}{2 \sigma^2} \right).
\end{align*}


\subsection{Discussions}
\label{section:Discuss}
We have proposed to minimize SCE for dimensionality reduction:
\begin{align*}
\mathrm{SCE}(\bY|\bZ) = -\frac{1}{2} \iint \left( \frac{\pzy}{\pz} - 1 \right)^2 \pz \dz \dy.
\end{align*}
On the other hand, in the previous work \citep{NC:Suzuki+Sugiyama:2013},
\emph{squared-loss mutual information} (SMI) 
was maximized for dimensionality reduction:
\begin{align*}
  \mathrm{SMI}(\bY,\bZ)
  = \frac{1}{2} \iint \left( \frac{\pzy}{\pz\py} - 1 \right)^2 \pz\py \dz \dy.
\end{align*}
This shows that the essential difference is whether $\py$
is included in the denominator of the density ratio.
Thus, if $\py$ is uniform, the proposed dimensionality reduction method
using SCE is reduced to the existing method using SMI.
However, if $\py$ is not uniform,
the density ratio function $\frac{\pzy}{\pz\py}$ included in SMI
may be more fluctuated than $\frac{\pzy}{\pz}$ included in SCE.
Since a smoother function can be more accurately estimated from a small number of samples
in general,
the proposed method using SCE is expected to work better
than the existing method using SMI.
We will experimentally demonstrate this effect in Section~\ref{chap:exp}.

\section{Experiments} \label{chap:exp}
In this section,
we experimentally investigate the practical usefulness of the proposed method.

\subsection{Illustration}
We consider the following dimensionality reduction schemes:
\begin{description}
  \item[None:] No dimensionality reduction is performed.
  \item[LSMI:]
    Dimension reduction is performed by
    maximizing an SMI approximator called \emph{least-squares MI} (LSMI)
    using natural gradients over the Grassmann manifold \citep{NC:Suzuki+Sugiyama:2013}.
  \item[LSCE (proposed):]
    Dimension reduction is performed by
    minimizing the proposed LSCE using natural gradients over the Grassmann manifold.
  \item[True (reference)]
    The ``true'' subspace is used
    (only for artificial data).
\end{description}
After dimension reduction, 
we execute the following conditional density estimators:
\begin{description}
  \item[$\epsilon$-KDE:]
    $\epsilon$-neighbor kernel density estimation,
    where $\epsilon$ is chosen by least-squares cross-validation.
  \item[LSCDE:]
    Least-squares conditional density estimation
    \citep{AISTATS:Sugiyama+etal:2010}.
\end{description}
Note that the proposed method,
which is the combination of LSCE and LSCDE,
does not explicitly require the post-LSCDE step
because LSCDE is executed inside LSCE.

\begin{figure*}[t]
  \centering
  \subfigure[Illustrative data]{
    \includegraphics[width=0.48\textwidth]{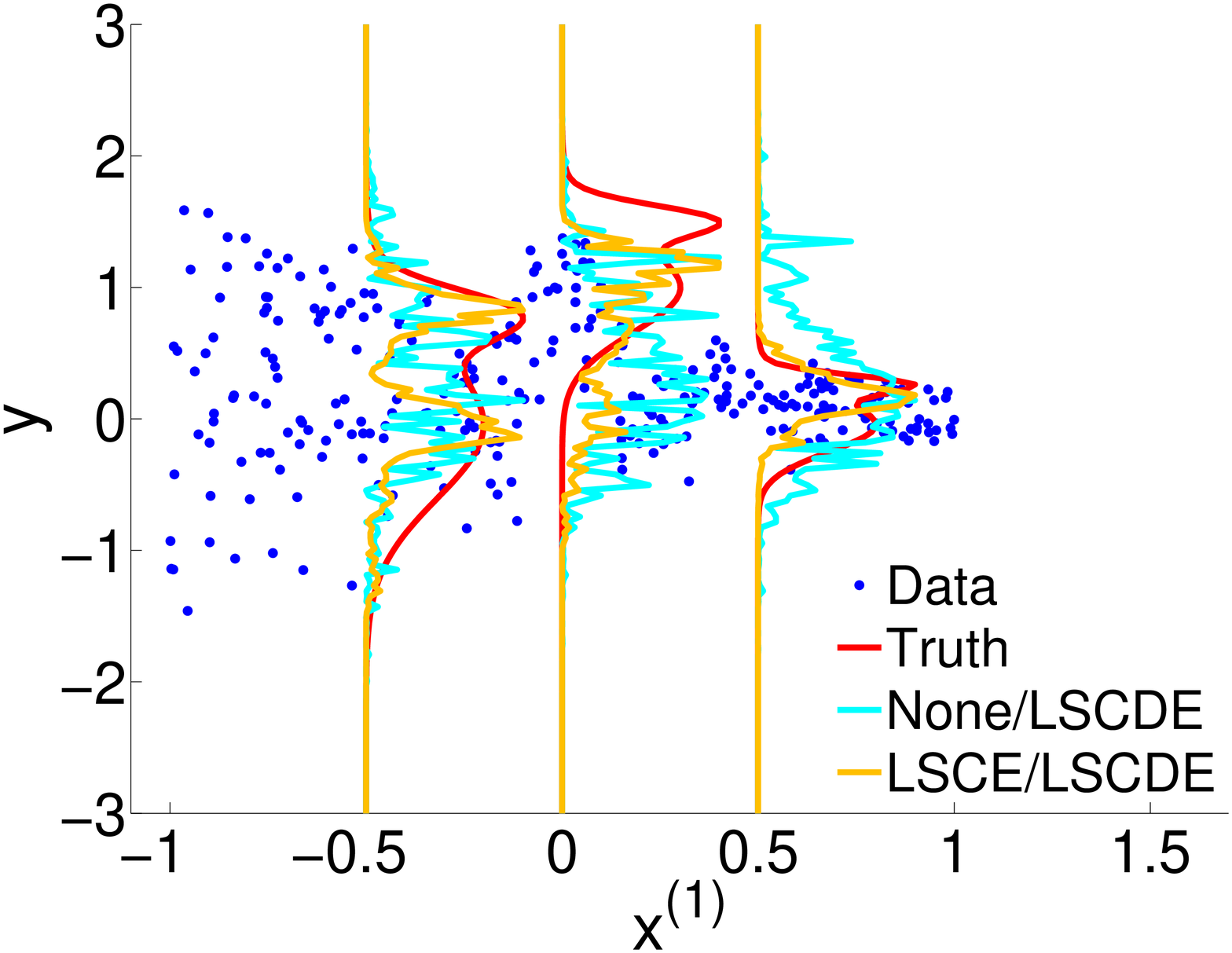}
  }
  \subfigure[Bone mineral density]{
    \includegraphics[width=0.48\textwidth]{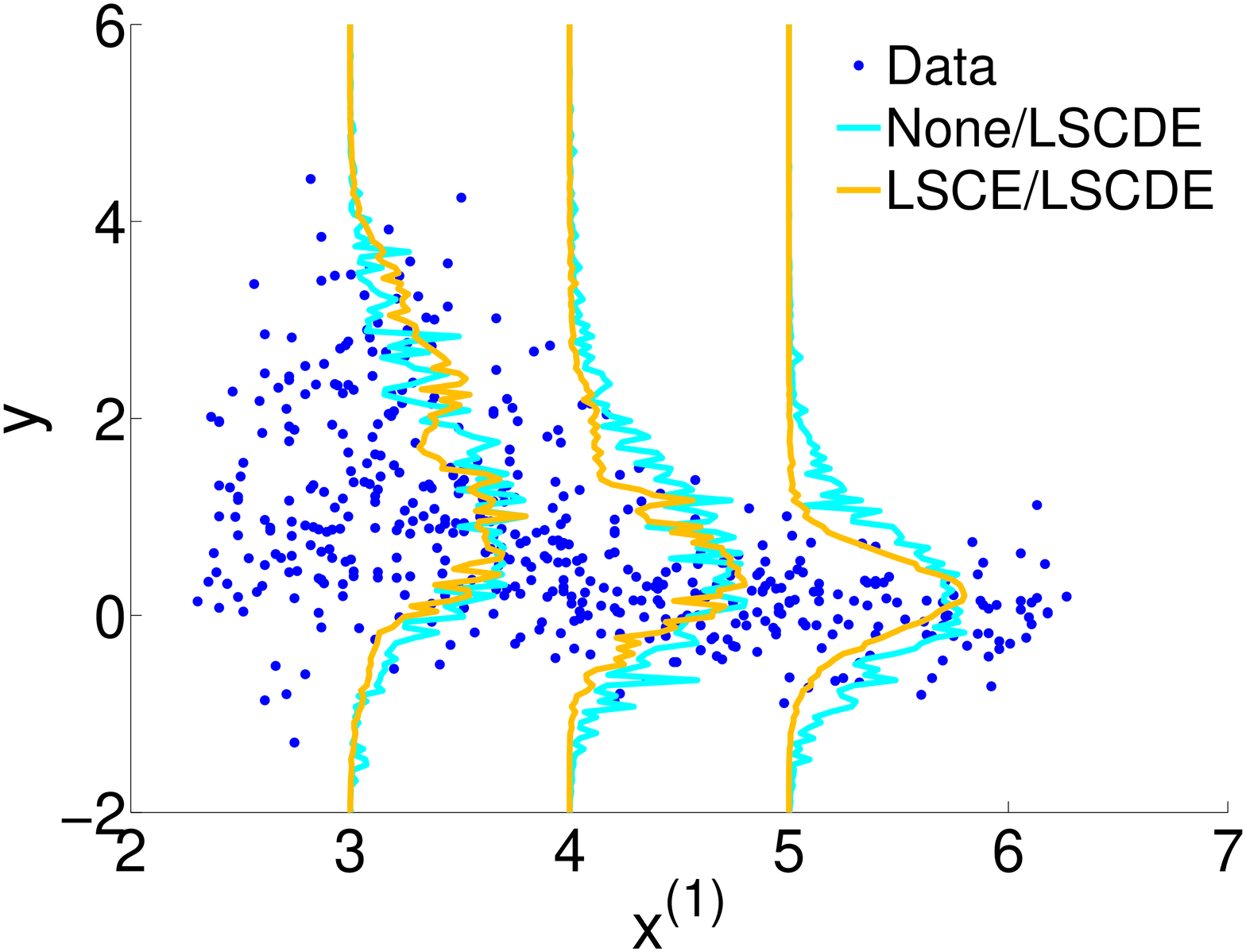}
  }
  \subfigure[Old faithful geyser]{
    \includegraphics[width=0.48\textwidth]{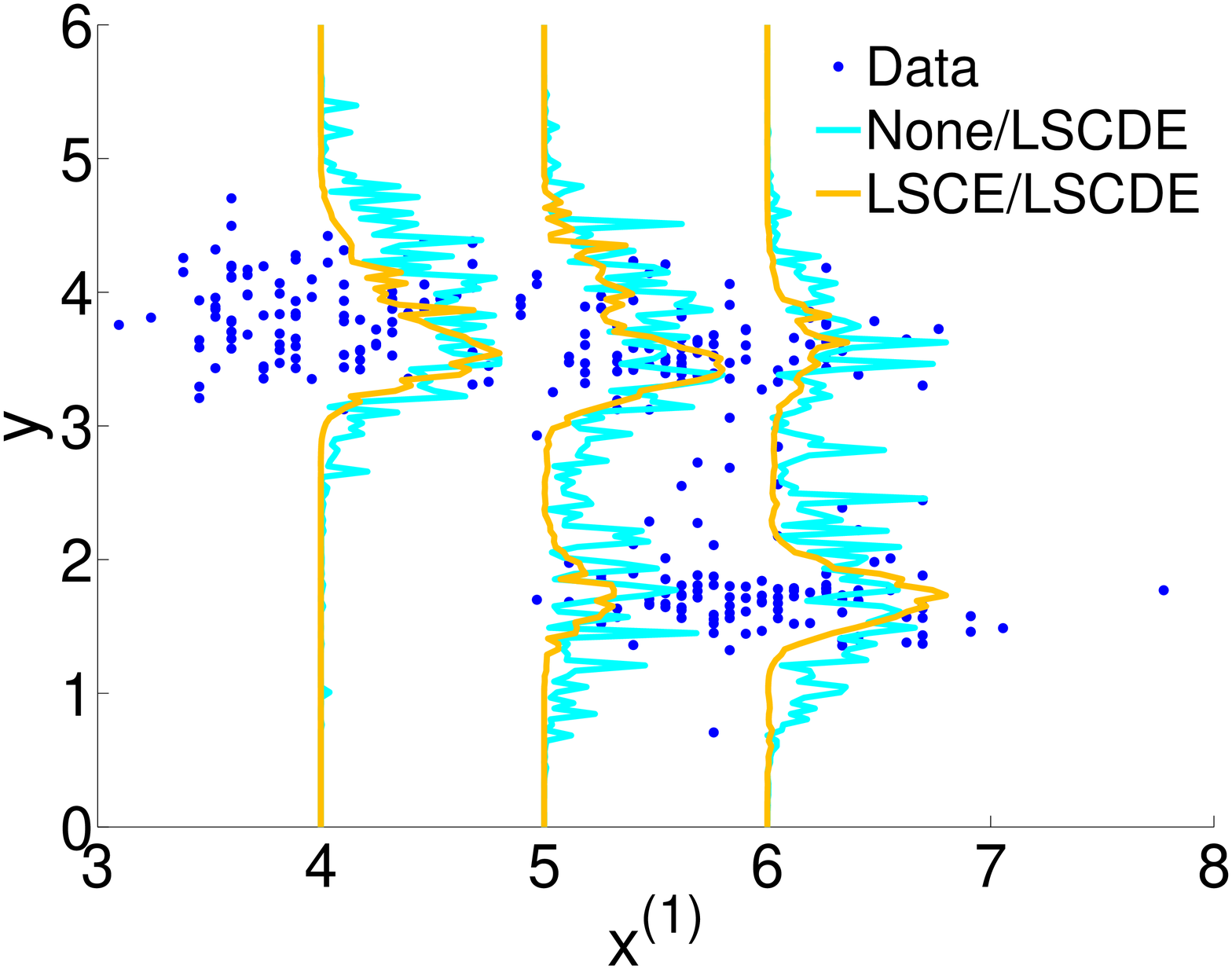}
  }
  \caption{Examples of conditional density estimation by
    plain LSCDE (None/LSCDE)
    and the proposed method (LSCE/LSCDE).
  }
    \label{fig:illustration}
\end{figure*}

First, we illustrate the behavior of the plain LSCDE (None/LSCDE)
and the proposed method (LSCE/LSCDE).
The datasets illustrated in Figure~\ref{fig:illustration}
have $d_\textbf{x} = 5$, $d_\textbf{y} = 1$, and $d_\textbf{z} = 1$.
The first dimension of input $\bx$ and output $y$ of the samples are plotted
in the graphs,
and other 4 dimensions of $\bx$ are just standard normal noise.
The results show that the plain LSCDE does not perform well
due to the irrelevant noise dimensions of $\bx$,
while the proposed method gives much better estimates.

\subsection{Artificial Datasets}
For $d_\textbf{x} = 5$, $d_\textbf{y} = 1$,
$\bx \sim \mathcal{N}(\bx|\boldsymbol{0}, \boldsymbol{I}_5)$, and $\epsilon \sim \mathcal{N}(\epsilon|0, 0.25^2)$,
where $\mathcal{N}(\cdot|\boldsymbol{\mu},\boldsymbol{\Sigma})$ denotes the normal distribution
with mean $\boldsymbol{\mu}$ and covariance matrix $\boldsymbol{\Sigma}$,
we consider the following artificial datasets:
\begin{description}
\item[(a)]
$d_\textbf{z} = 2$ and $y = (x^{(1)})^2 + (x^{(2)})^2 + \epsilon$.
\item[(b)]
$d_\textbf{z} = 1$ and $y = x^{(2)} + (x^{(2)})^2 + (x^{(2)})^3 + \epsilon$.
\end{description}

\begin{figure*}[t]
  \centering
  \subfigure[Artificial data 1]{
    \includegraphics[width=0.31\textwidth]{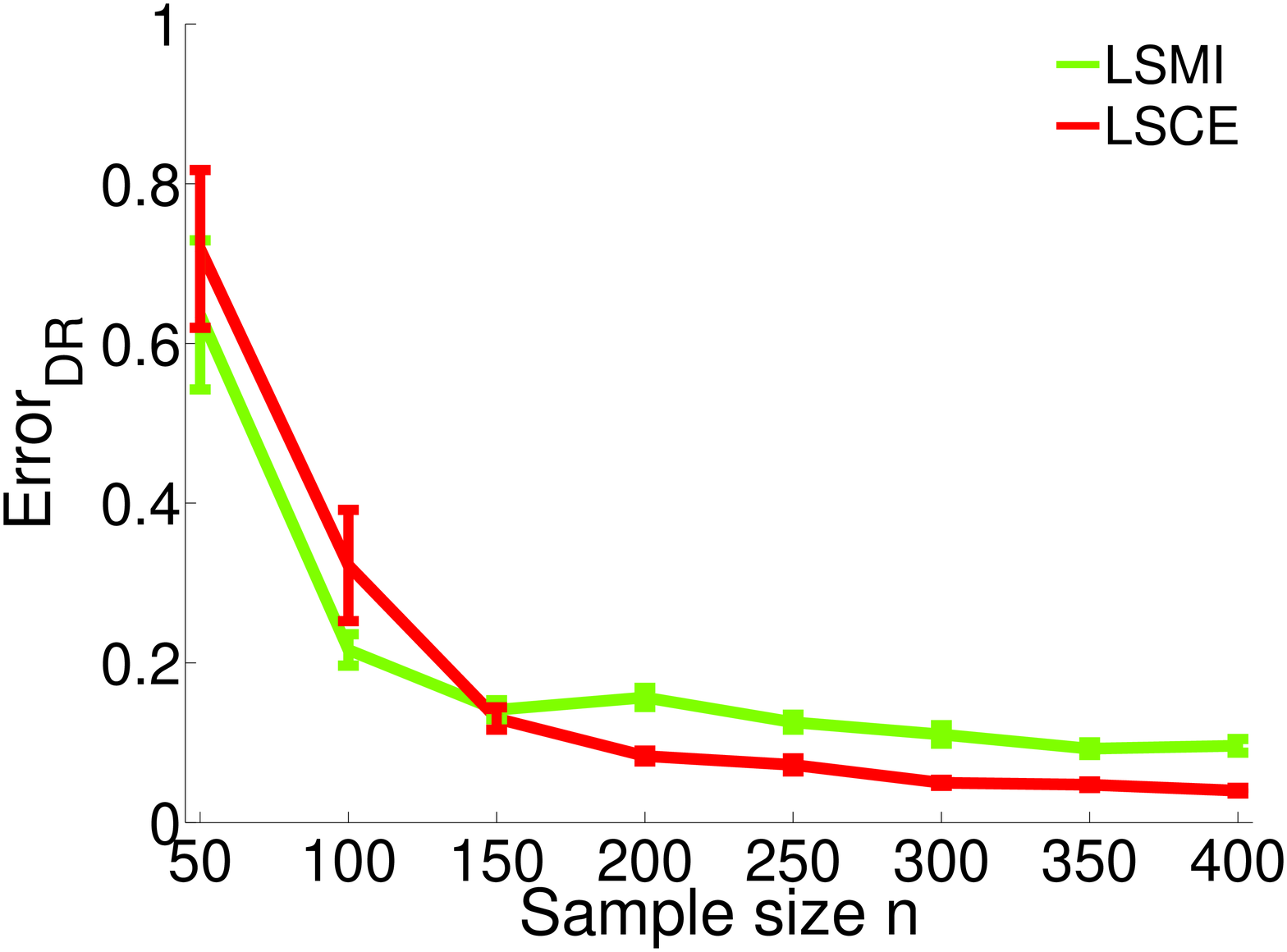}
    \includegraphics[width=0.31\textwidth]{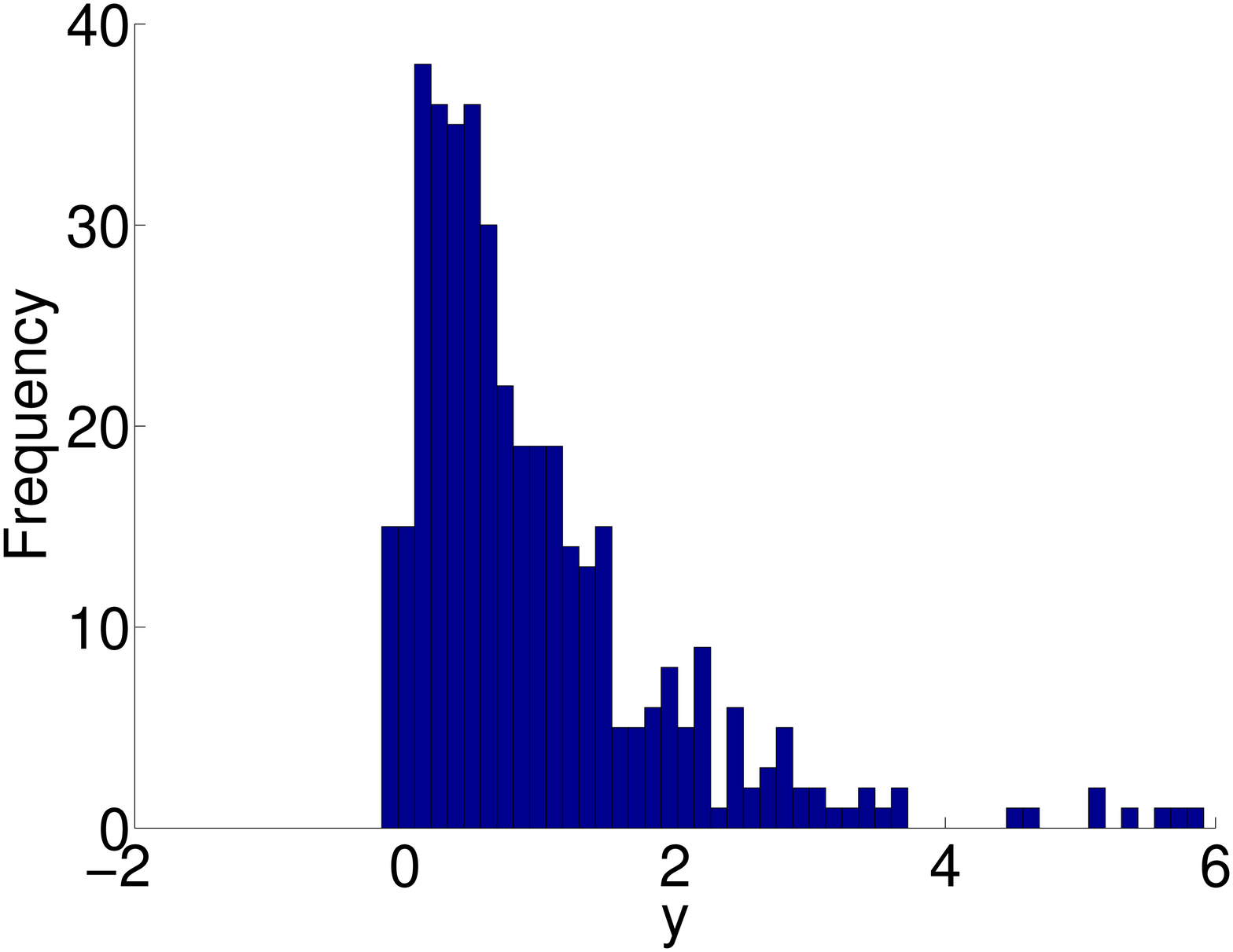}
    \includegraphics[width=0.31\textwidth]{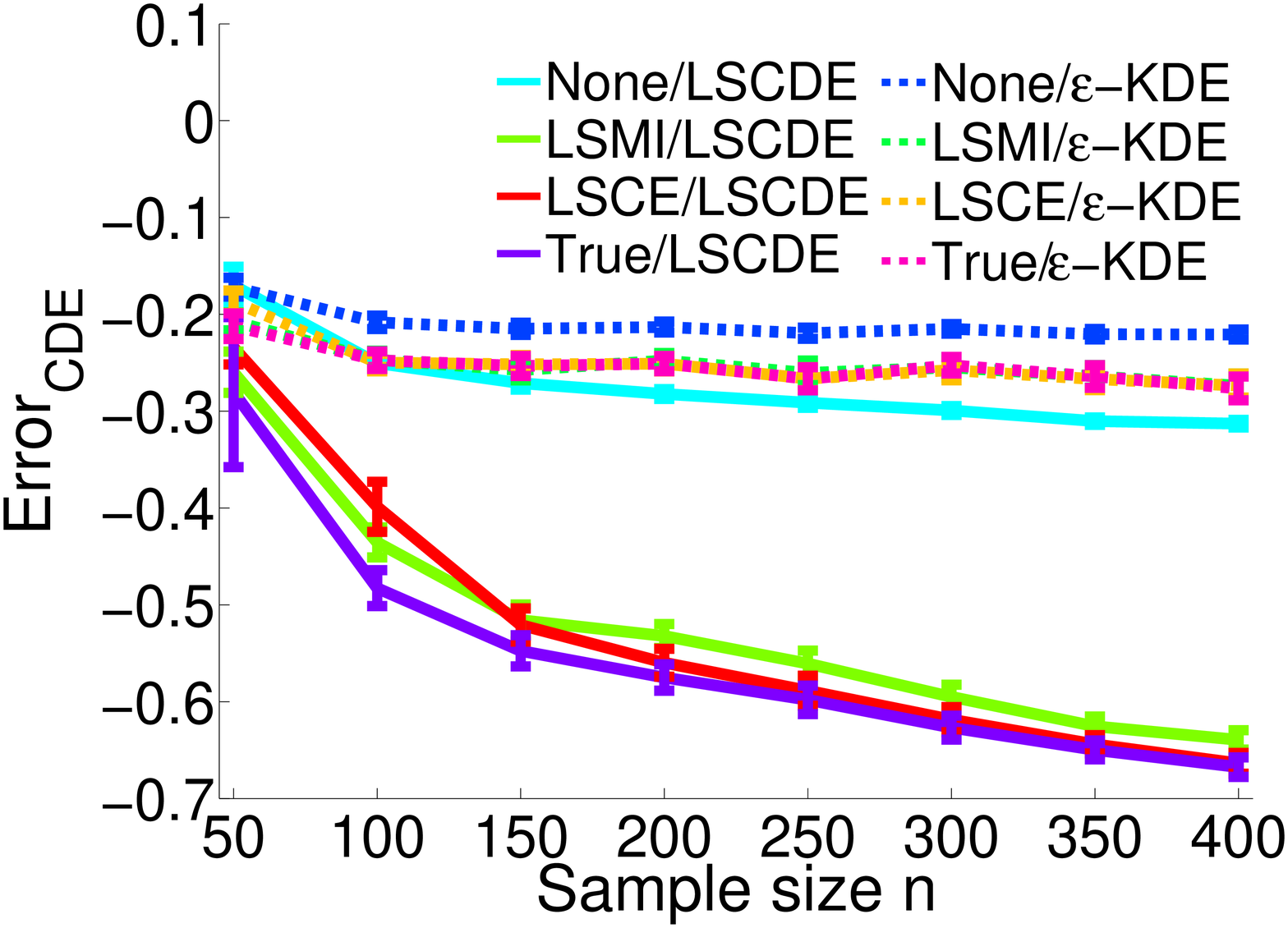}
    \label{fig:toy3_dr_cde}
  }\\
  \subfigure[Artificial data 2]{
    \includegraphics[width=0.31\textwidth]{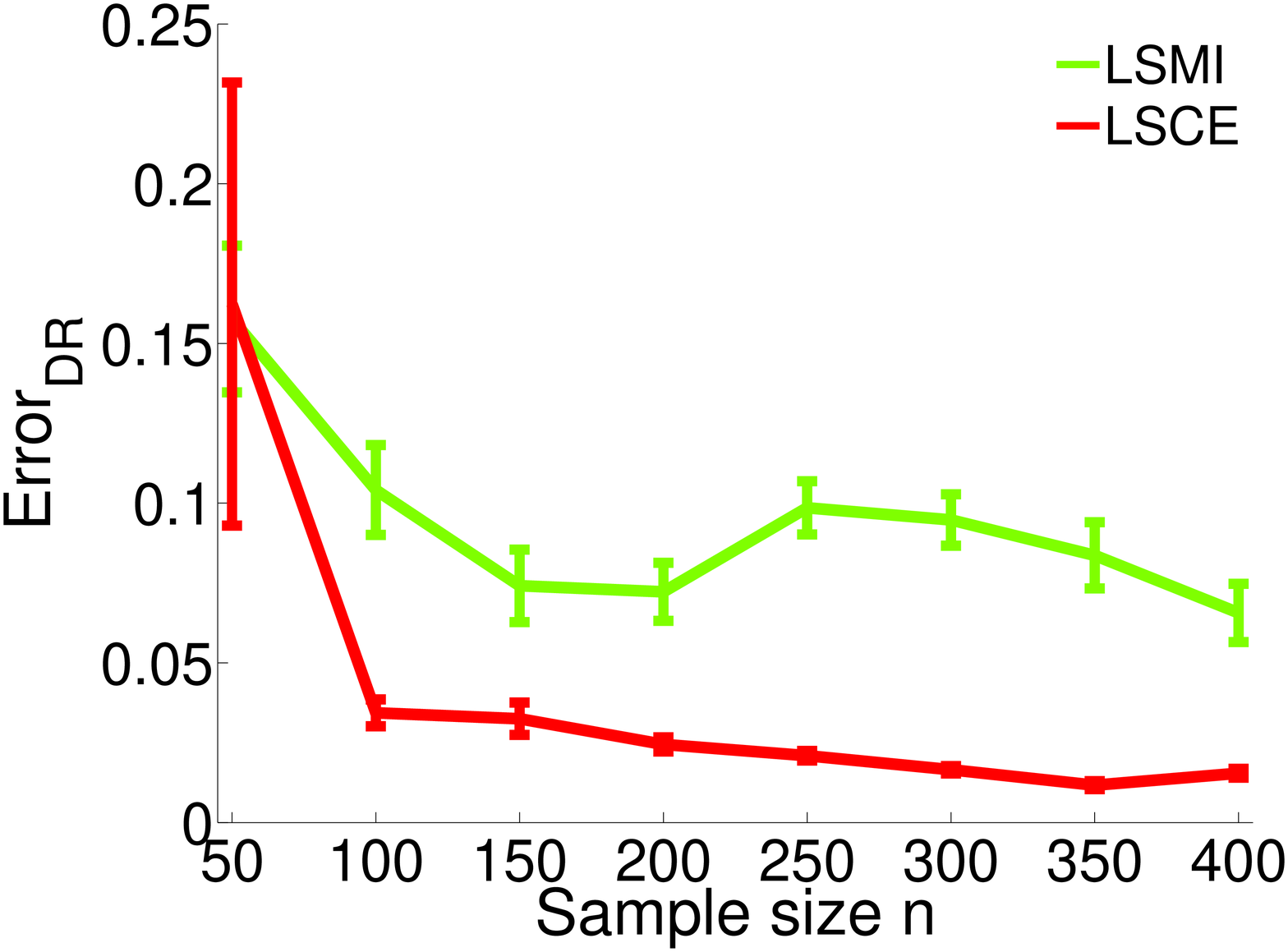}
    \includegraphics[width=0.31\textwidth]{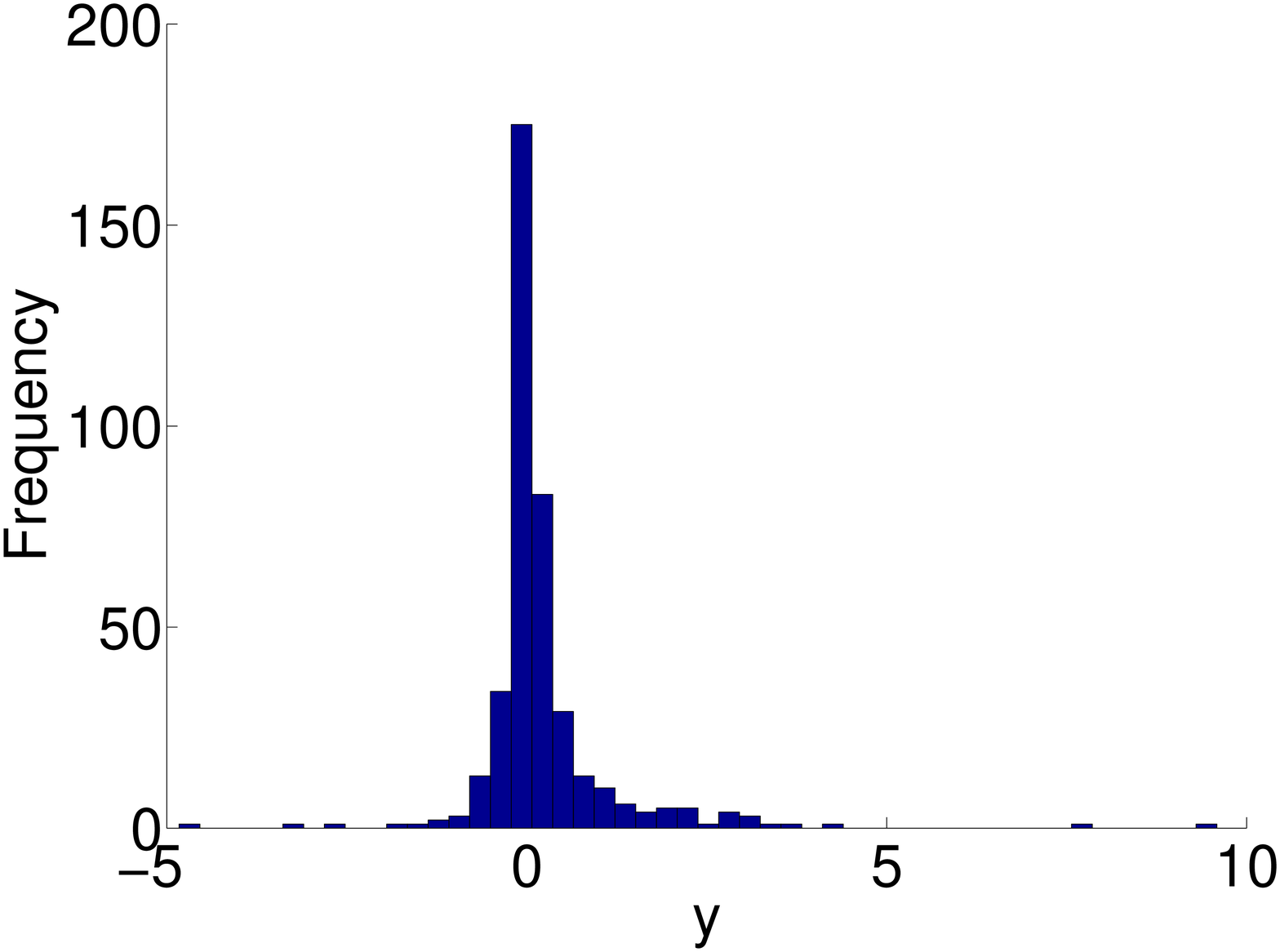}
    \includegraphics[width=0.31\textwidth]{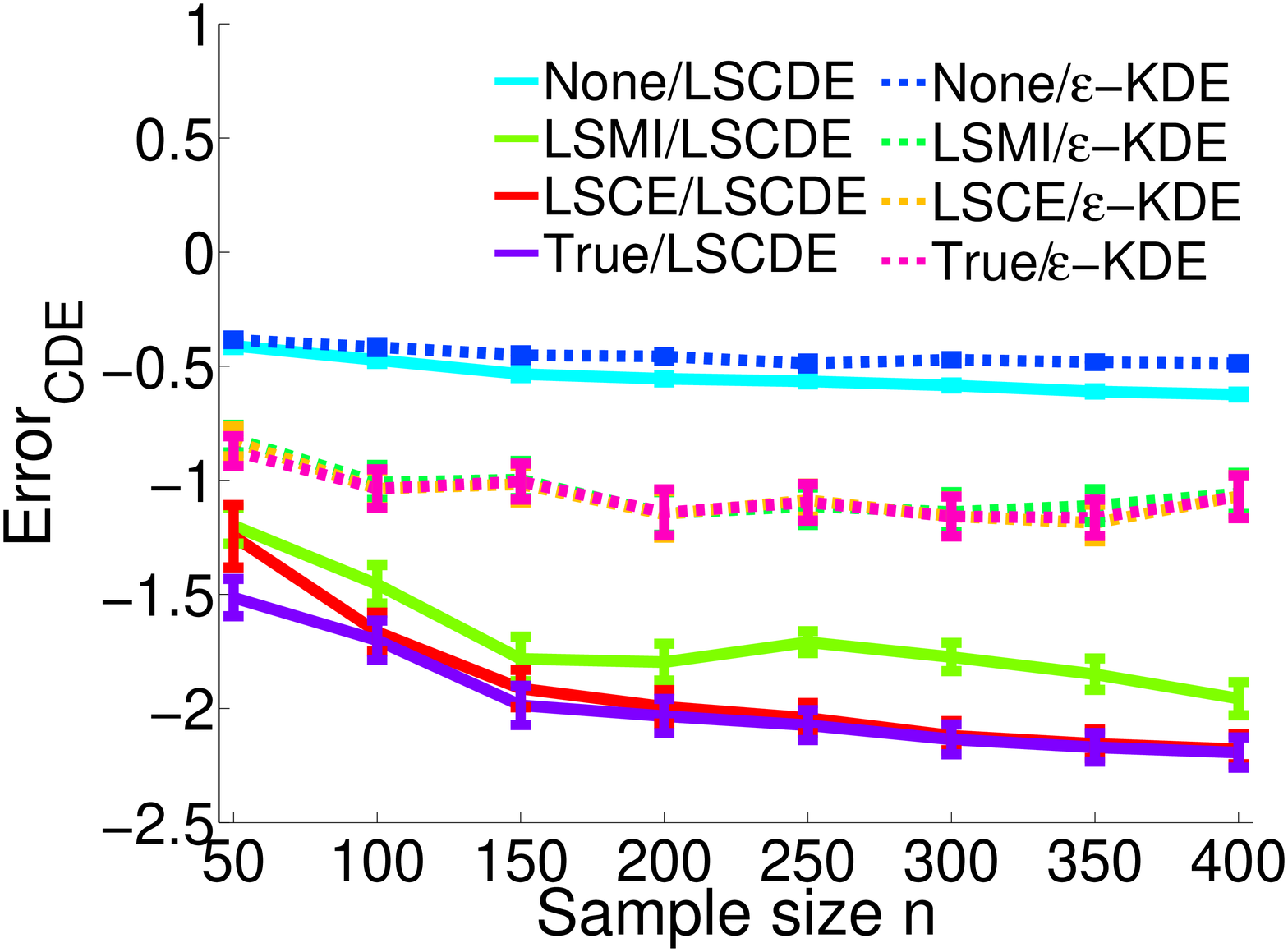}
  \label{fig:toy2_dr_cde}
  }
  \caption{Left column: The mean and standard error of the dimensionality reduction error over 20 runs
    on the artificial datasets.
    Middle column: Histograms of $\{y_i\}_{i=1}^{400}$.
    Right column: The mean and standard error of the conditional density estimation error over 20 runs.
  }
    \label{fig:toy_dr_cde}
\end{figure*}

The left column of Figure~\ref{fig:toy_dr_cde} shows the dimensionality reduction error
between true $\bW^*$ and its estimate $\widehat{\bW}$ 
for different sample size $n$, measured by
\begin{align*}
\mathrm{Error}_{\mathrm{DR}} = \| \widehat{\bW}^\top \bhW - {\bW^*}^\top\bW^* \|_{\mathrm{Frobenius}},
\end{align*}
where $\|\cdot\|_{\mathrm{Frobenius}}$ denotes the Frobenius norm.
LSMI and LSCE perform similarly for the dataset (a),
while LSCE clearly outperforms LSMI for the datasets (b).
To explain this difference,
we plot the histograms of $\{y\}_{i=1}^{400}$ in the middle column of Figure~\ref{fig:toy_dr_cde}.
They show that the profile of the histogram (which is a sample approximation of $p(y)$)
in the dataset (b) is much sharper than that in the dataset (a).
As discussed in Section~\ref{section:Discuss},
the density ratio $\frac{\pzy}{\pz \py}$ used in LSMI contains $\py$.
Thus, for the dataset (b), the density ratio $\frac{\pzy}{\pz \py}$ would be highly non-smooth
and thus is hard to approximate.
On the other hand,
the density ratio used in SCE is $\frac{\pzy}{\pz}$,
where $\py$ is not included.
Therefore, $\frac{\pzy}{\pz}$ would be smoother than $\frac{\pzy}{\pz \py}$
and $\frac{\pzy}{\pz}$ is easier to estimate than $\frac{\pzy}{\pz \py}$.

The right column of Figure~\ref{fig:toy_dr_cde} plots
the conditional density estimation error between true $p(\by|\bx)$ 
and its estimate $\widehat{p}(\by|\bx)$, evaluated by the squared-loss:
\begin{align*}
\mathrm{Error}_{\mathrm{CDE}} = \frac{1}{2n'} \sum_{i=1}^{n'}
\int \widehat{p}(\by|\widetilde{\bx}_i)^2 \dy 
- \frac{1}{n'} \sum_{i=1}^{n'} \widehat{p}(\widetilde{\by}_i|\widetilde{\bx}_i),
\end{align*}
where $\{(\widetilde{\bx}_i, \widetilde{\by}_i)\}_{i=1}^{n'}$ is a set of test samples
that have not been used for training. We set $n' = 1000$.
The graphs show that LSCDE overall outperforms $\epsilon$-KDE for both datasets.
For the dataset (a), LSMI/LSCDE and LSCE/LSCDE perform equally well,
which are much better than no dimension reduction (None/LSCDE)
and are comparable to the method with the true subspace (True/LSCDE).
For the dataset (b), LSCE/LSCDE outperforms LSMI/LSCDE and None/LSCDE,
and is comparable to the method with the true subspace (True/LSCDE).

\subsection{Benchmark Datasets}
Next, we use the UCI benchmark datasets \citep{Bache+Lichman:2013}. 
We randomly select $n$ samples from each dataset for training,
and the rest are used to measure the conditional density estimation error in the test phase.
Since the dimensionality of the subspace $d_\textbf{z}$ is unknown,
we chose it by cross-validation.
The results are summarized in Table~\ref{table:benchrobot_cde},
showing that that the proposed method, LSCE/LSCDE works well overall.
Table~\ref{table:benchrobot_dim} describes the dimensionalities selected by cross-validation,
showing that both LSCE and LSMI reduce the dimensionalty significantly.
For ``Housing'', ``AutoMPG'', ``Energy'', and ``Stock'', 
LSMI/LSCDE tends to more aggressively reduce the dimensionality than LSCE/LSCDE.

\begin{sidewaystable}
	\centering
\caption{Mean and standard error of the conditional density estimation error over 10 runs.
The best method in term of the mean error and comparable methods according to the two-sample paired \textit{t-test} 
at the significance level $5\%$ are specified by bold face.
}
\label{table:benchrobot_cde}
\vspace*{2mm}
\begin{tabular}{@{} c @{\ }|@{\ } c@{\ } c @{\ }|c @{\ }|@{\ }c|c@{\ }|@{\ }c|c@{\ }|@{\ }c | c@{} }  
\hline   
\multirow{2}{*}{Dataset} & \multirow{2}{*}{$(d_\textbf{x}, d_\textbf{y})$} & \multirow{2}{*}{$n$} & \multicolumn{2}{|c|}{LSCE} & \multicolumn{2}{|c|}{LSMI} & \multicolumn{2}{|c|}{No reduction} & \multirow{2}{*}{Scale} \\ \cline{4-9} 
 & & & LSCDE & $\epsilon$-KDE & LSCDE & $\epsilon$-KDE & LSCDE & $\epsilon$-KDE & \\
\hline
Housing & $(13,1)$ & 100 		 
		 & $\boldsymbol{-1.73(0.09)}$
         & $-1.57(0.11)$
         & $\boldsymbol{-1.91(0.05)}$
         & $-1.62(0.08)$
         & $-1.41(0.05)$
         & $-1.13(0.01)$ 
         & $\times 1$ \\
Auto MPG & $(7,1)$ & 100		 
		 & $\boldsymbol{-1.80(0.04)}$
         & $\boldsymbol{-1.74(0.06)}$
         & $\boldsymbol{-1.85(0.04)}$
         & $\boldsymbol{-1.77(0.05)}$
         & $\boldsymbol{-1.75(0.04)}$
         & $-1.46(0.04)$ 
         & $\times 1$ \\   
Servo & $(4,1)$ & 50		 
		 & $\boldsymbol{-2.92(0.18)}$
         & $\boldsymbol{-3.03(0.14)}$
         & $\boldsymbol{-2.69(0.18)}$
         & $\boldsymbol{-2.95(0.11)}$
         & $-2.62(0.09)$
         & $-2.72(0.06)$ 
         & $\times 1$ \\ 
Yacht & $(6,1)$ & 80		 
		 & $\boldsymbol{-6.46(0.02)}$
         & $\boldsymbol{-6.23(0.14)}$
         & $-5.63(0.26)$
         & $-5.47(0.29)$
         & $-1.72(0.04)$
         & $-2.95(0.02)$ 
         & $\times 1$ \\ 
Physicochem & $(9,1)$ & 500		 
		 & $\boldsymbol{-1.19(0.01)}$
         & $-0.99(0.02)$
         & $\boldsymbol{-1.20(0.01)}$
         & $-0.97(0.02)$
         & $\boldsymbol{-1.19(0.01)}$
         & $-0.91(0.01)$ 
         & $\times 1$ \\ 
White Wine & $(11,1)$ & 400 
		 & $-2.31(0.01)$
         & $\boldsymbol{-2.47(0.15)}$
         & $\boldsymbol{-2.35(0.02)}$
         & $\boldsymbol{-2.60(0.12)}$
         & $-2.06(0.01)$
         & $-1.89(0.01)$ 
         & $\times 1$ \\ 
Red Wine & $(11,1)$ & 300 
		 & $\boldsymbol{-2.85(0.02)}$
         & $-1.95(0.17)$
         & $\boldsymbol{-2.82(0.03)}$
         & $-1.93(0.17)$
         & $-2.03(0.02)$
         & $-1.13(0.04)$ 
         & $\times 1$ \\ 
Forest Fires & $(12,1)$ & 100 		 
		 & $\boldsymbol{-7.18(0.02)}$
         & $-6.93(0.03)$
         & $-6.93(0.04)$
         & $-6.93(0.02)$
         & $-3.40(0.07)$
         & $-6.96(0.02)$ 
         & $\times 1$ \\   
Concrete & $(8,1)$ & 300		 
		 & $\boldsymbol{-1.36(0.03)}$
         & $-1.20(0.06)$
         & $\boldsymbol{-1.30(0.03)}$
         & $-1.18(0.04)$
         & $-1.11(0.02)$
         & $-0.80(0.03)$ 
         & $\times 1$ \\ 
Energy & $(8,2)$ & 200	 
		 & $\boldsymbol{-7.13(0.04)}$
         & $-4.18(0.22)$
         & $-6.04(0.47)$
         & $-3.41(0.49)$
         & $-2.12(0.06)$
         & $-1.95(0.14)$ 
         & $\times 10$ \\ 
Stock & $(7,2)$ & 100		 
		 & $-8.37(0.53)$
         & $\boldsymbol{-9.75(0.37)}$
         & $\boldsymbol{-9.42(0.50)}$
         & $\boldsymbol{-10.27(0.33)}$
         & $-7.35(0.13)$
         & $-9.25(0.14)$ 
         & $\times 1$ \\ 
         \hline 
2 Joints & $(6,4)$ & 100 
		 & $\boldsymbol{-10.49(0.86)}$ 
         & $-7.50(0.54)$  
         & $\boldsymbol{-8.00(0.84)}$ 
         & $-7.44(0.60)$ 
         & $-3.95(0.13)$ 
         & $-3.65(0.14)$ 
         & $\times 1$ \\ 
4 Joints & $(12,8)$ & 200 
		 & $\boldsymbol{-2.81(0.21)}$ 
         & $-1.73(0.14)$  
         & $-2.06(0.25)$ 
         & $-1.38(0.16)$ 
         & $-0.83(0.03)$ 
         & $-0.75(0.01)$ 
         & $\times 10$ \\ 
9 Joints & $(27,18)$ & 500 
		 & $\boldsymbol{-8.37(0.83)}$ 
         & $-2.44(0.17)$  
         & $\boldsymbol{-9.74(0.63)}$ 
         & $-2.37(0.51)$ 
         & $-1.60(0.36)$ 
         & $-0.89(0.02)$ 
         & $\times 100$ \\  \hline
Sumi-e 1 & $(9,6)$ & 200 
		 & $\boldsymbol{-9.96(1.60)}$
		 & $-1.49(0.78)$
		 & $\boldsymbol{-6.00(1.28)}$
		 & $1.24(1.99)$
		 & $-5.98(0.80)$
		 & $-0.17(0.44)$
         & $\times 10$ \\ 
Sumi-e 2 & $(9,6)$ & 250 
		 & $\boldsymbol{-16.83(1.70)}$
		 & $-2.22(0.97)$
		 & $-9.54(1.31)$
		 & $-3.12(0.75)$
		 & $-7.69(0.62)$
		 & $-0.66(0.13)$
         & $\times 10$ \\ 
Sumi-e 3 & $(9,6)$ & 300 
		 & $\boldsymbol{-24.92(1.92)}$
		 & $-6.61(1.25)$
		 & $-18.0(2.61)$
		 & $-4.47(0.68)$
		 & $-8.98(0.66)$
		 & $-1.45(0.43)$
         & $\times 10$ \\ \hline
\end{tabular}
\end{sidewaystable}

\begin{table}[t]
	\centering
	\caption{
	Mean and standard error of the chosen dimensionality over 10 runs.
	}
\label{table:benchrobot_dim}
\vspace*{2mm}
\begin{tabular}{@{} c | c | c | c | c | c @{} }  
\hline   
\multirow{2}{*}{Data set} & \multirow{2}{*}{$(d_\textbf{x}, d_\textbf{y})$} & 
\multicolumn{2}{|c|}{LSCE} & \multicolumn{2}{|c}{LSMI} \\ \cline{3-6}
 & & LSCDE & $\epsilon$-KDE  & LSCDE & $\epsilon$-KDE \\ 
\hline
Housing & $(13,1)$  		 
         & $3.9(0.74)$ 
         & $2.0(0.79)$
         & $2.0(0.39)$ 
         & $1.3(0.15)$\\ 
Auto MPG & $(7,1)$ 		 
         & $3.2(0.66)$ 
         & $1.3(0.15)$
         & $2.1(0.67)$ 
         & $1.1(0.10)$\\ 
Servo & $(4,1)$ 		 
         & $1.9(0.35)$ 
         & $2.4(0.40)$
         & $2.2(0.33)$ 
         & $1.6(0.31)$ \\ 
Yacht & $(6,1)$ 		 
         & $1.0(0.00)$ 
         & $1.0(0.00)$
         & $1.0(0.00)$ 
         & $1.0(0.00)$ \\ 
Physicochem & $(9,1)$ 	 
         & $6.5(0.58)$ 
         & $1.9(0.28)$
         & $6.6(0.58)$ 
         & $2.6(0.86)$ \\ 
White Wine & $(11,1)$  
         & $1.2(0.13)$ 
         & $1.0(0.00)$
         & $1.4(0.31)$ 
         & $1.0(0.00)$ \\ 
Red Wine & $(11,1)$  
         & $1.0(0.00)$ 
         & $1.3(0.15)$
         & $1.2(0.20)$ 
         & $1.0(0.00)$ \\ 
Forest Fires & $(12,1)$  		 
         & $1.2(0.20)$ 
         & $4.9(0.99)$
         & $1.4(0.22)$
         & $6.8(1.23)$ \\ 
Concrete & $(8,1)$ 		 
         & $1.0(0.00)$ 
         & $1.0(0.00)$
         & $1.2(0.13)$ 
         & $1.0(0.00)$ \\ 
Energy & $(8,2)$ 	 
         & $5.9(0.10)$ 
         & $3.9(0.80)$
         & $2.1(0.10)$ 
         & $2.0(0.30)$ \\ 
Stock & $(7,2)$ 		 
         & $3.2(0.83)$ 
         & $2.1(0.59)$
         & $2.1(0.60)$ 
         & $2.7(0.67)$ \\ 
         \hline 
2 Joints & $(6,4)$  
         & $2.9(0.31)$ 
         & $2.7(0.21)$
         & $2.5(0.31)$ 
         & $2.0(0.00)$ \\ 
4 Joints & $(12,8)$  
         & $5.2(0.68)$ 
         & $6.2(0.63)$
         & $5.4(0.67)$ 
         & $4.6(0.43)$ \\ 
9 Joints & $(27,18)$  
         & $13.8(1.28)$ 
         & $15.3(0.94)$
         & $11.4(0.75)$ 
         & $13.2(1.02)$ \\   \hline
Sumi-e 1 & $(9,6)$  
		 & $5.3(0.72)$
		 & $2.9(0.85)$
		 & $4.5(0.45)$
		 & $3.2(0.76)$ \\
Sumi-e 2 & $(9,6)$  
		 & $4.2(0.55)$
		 & $4.4(0.85)$
		 & $4.6(0.87)$
		 & $2.5(0.78)$ \\
Sumi-e 3 & $(9,6)$  
		 & $3.6(0.50)$
		 & $2.7(0.76)$
		 & $2.6(0.40)$
		 & $1.6(0.27)$ \\ \hline
\end{tabular}
\end{table}

\begin{figure}[t]
	\centering
        \includegraphics[width=0.47\textwidth]{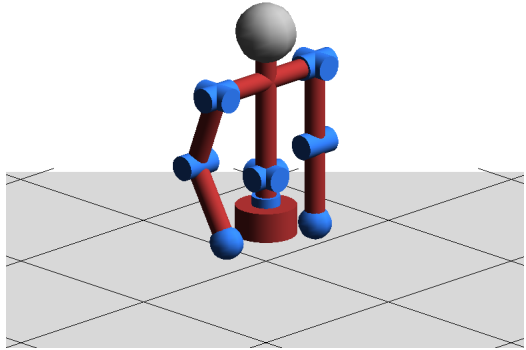}
	\caption{Simulator of the upper-body part of the humanoid robot \emph{CB-i}.}
	\label{fig:robot}
\end{figure}

\subsection{Humanoid Robot}
We evaluate the performance of the proposed method on humanoid robot transition estimation.
We use a simulator of the upper-body part of the humanoid robot \emph{CB-i} \citep{Cheng2007}
(see Figure~\ref{fig:robot}).
The robot has 9 controllable joints: shoulder pitch, shoulder roll, elbow pitch of the right arm, shoulder pitch,
shoulder roll, elbow pitch of the left arm, waist yaw, torso roll, and torso pitch joints.

Posture of the robot is described by 18-dimensional real-valued state vector $\bs$,
which corresponds to the angle and angular velocity of each joint in radians and radians per seconds, respectively. 
We can control the robot by sending the action command $\ba$ to the system.
The action command $\ba$ is a 9-dimensional real-valued vector, which corresponds to the target angle of each joint.
When the robot is currently at state $\bs$ and receives action $\ba$,
the physical control system of the simulator calculates the amount of torques
to be applied to each joint. These torques are calculated by the \emph{proportional-derivative} (PD) controller as
\begin{align*}
\tau_i = K_{p_i}(a_i - s_i) - K_{d_i}\dot{s}_i,
\end{align*}
where $s_i$, $\dot{s}_i$, and $a_i$ denote the current angle, the current angular velocity, and the received target angle of the $i$-th joint, respectively. 
$K_{p_i}$ and $K_{d_i}$ denote the position and velocity gains for the $i$-th joint, respectively. 
We set $K_{p_i} = 2000$ and $K_{d_i} = 100$ for all joints except that $K_{p_i} = 200$ and $K_{d_i} = 10$ for the elbow pitch joints.
After the torques are applied to the joints, the physical control system update the state of the robot to $\bs'$.

In the experiment, we randomly choose the action vector $\ba$ and simulate a noisy control system
by adding a bimodal Gaussian noise vector.
More specifically, the action $a_i$ of the $i$-th joint is first drawn from uniform distribution on 
$[s_i-0.087, s_i+0.087]$. The drawn action is then contaminated by Gaussian noise with mean 0 and standard deviation 0.034
with probability 0.6 and Gaussian noise with mean -0.087 and standard deviation 0.034 with probability 0.4. 
By repeatedly control the robot $n$ times, we obtain the transition samples $\{(\bs_j, \ba_j, \bs'_{j})\}_{j=1}^n$.
Our goal is to learn the system dynamic as a state transition probability $p(\bs'|\bs, \ba)$ from these samples.
Thus, as the conditional density estimation problem,
the state-action pair $(\bs^\top, \ba^\top)^\top$ is regarded as input variable $\bx$,
while the next state $\bs'$ is regarded as output variable $\by$.
Such state-transition probabilities are highly useful in model-based reinforcement learning \citep{book:Sutton+Barto:1998}.

We consider three scenarios: Using only 2 joints (right shoulder pitch and right elbow pitch),
only 4 joints (in addition, right shoulder roll and waist yaw),
and all 9 joints.
Thus, $d_\textbf{x}=6$ and $d_\textbf{y}=4$ for the 2-joint case,
$d_\textbf{x}=12$ and $d_\textbf{y}=8$ for the 4-joint case,
and
$d_\textbf{x}=27$ and $d_\textbf{y}=18$ for the 9-joint case.
We generate $500$, $1000$, and $1500$ transition samples
for the 2-joint, 4-joint, and 9-joint cases.
We then randomly choose $n=100$, $200$, and $500$ samples for training,
and use the rest for evaluating the test error.
The results are summarized also in Table~\ref{table:benchrobot_cde},
showing that the proposed method performs well for the all three cases.
Table~\ref{table:benchrobot_dim} describes the dimensionalities selected by cross-validation,
showing that the humanoid robot's transition is highly redundant.



\begin{figure}[t]
  \centering
  \includegraphics[width=0.30\textwidth]{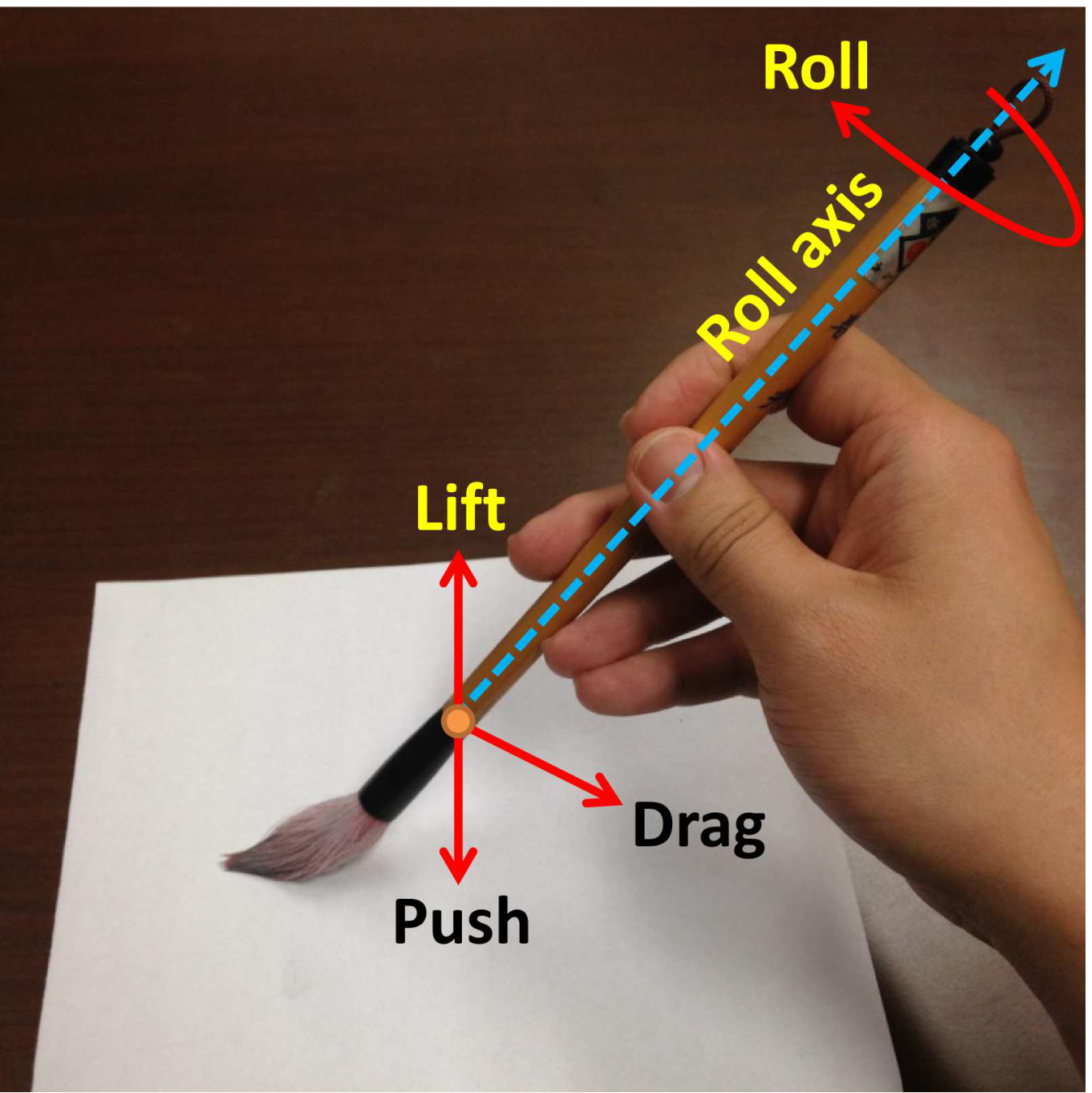}
  ~~~
\includegraphics[width=0.10\textwidth]{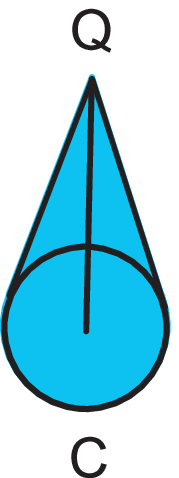}
\caption{Three actions of the brush, which is modeled as
  the footprint on a paper canvas.}
\label{fig:sumie}
\vspace*{-3mm}
\end{figure}

\subsection{Computer Art}

Finally, we consider the transition estimation 
problem in \emph{sumi-e} style brush drawings
for non-photorealistic rendering \citep{ICML:Xie+etal:2012}.
Our aim is to learn the brush dynamics as state transition
probability $p(\boldsymbol{s'}|\boldsymbol{s},\boldsymbol{a})$ from the real
artists' stroke-drawing samples. 

From a video of real brush stroks,
we extract footprints
and identify corresponding 3-dimensional actions
(see Figure~\ref{fig:sumie}).
The state vector consists of six measurements:
the angle of the velocity vector
and the heading direction of the footprint
relative to the medial axis of the drawing shape,
the ratio of the offset distance from the
center of the footprint to the nearest point on the medial axis over the
radius of the footprint, the relative curvatures of the nearest
current point and the next point on the medial axis, and
the binary signal of the reverse driving or not.
Thus, the state transition
probability $p(\boldsymbol{s'}|\boldsymbol{s},\boldsymbol{a})$
has $9$-dimensional input and $6$-dimensional output.
We collect $722$ transition samples in total.
We randomly choose $n = 200, 250$, and $300$ for training 
and use the rest for testing.


The estimation results summarized at the bottom of Table~\ref{table:benchrobot_cde} and 
Table~\ref{table:benchrobot_dim}.
These tables show that there exists a low-dimensional sufficient subspace and
the proposed method can successfully find it.

\section{Conclusion} \label{chap:conclusion} 
We proposed a new method for conditional density estimation in high-dimension problems. 
The key idea of the proposed method is to perform sufficient dimensionality reduction
by minimizing the square-loss conditional entropy (SCE),
which can be estimated by least-squares conditional density estimation.
Thus, dimensionality reduction and conditional density estimation are carried out simultaneously
in an integrated manner.
We have also shown that SCE and the squared-loss mutual information (SMI)
are similar but different in that the output density is included in the denominator of the density ratio in SMI.
This means that estimation of SMI is hard when the output density is fluctuated,
while the proposed method using SCE does not suffer from this problem.
The effectiveness of the proposed method was demonstrated through extensive experiments 
including humanoid robot transition and computer art.


\newpage

{\small
\bibliography{references,mybib,sugiyama}
\bibliographystyle{plainnat}
}


\newpage
\appendix
\onecolumn

\section{Proof of Theorem~\ref{theorem:SCE(Y|Z)-SCE(Y|X)>=0}}
\label{proof:SCE(Y|Z)-SCE(Y|X)>=0}
The $\widetilde{\mathrm{SCE}}$ is defined as
\begin{align*}
\widetilde{\mathrm{SCE}}(\bY|\bZ) &= -\frac{1}{2} \iint \pygz^2\pz\dz\dy.
\end{align*}
Then we have
\begin{align*}
\widetilde{\mathrm{SCE}}(\bY|\bZ) - \widetilde{\mathrm{SCE}}(\bY|\bX) &= \frac{1}{2}\iint  \pygx  ^2 \px \dy \dx 
							  - \frac{1}{2}\iint  \pygz  ^2 \pz \dz \dy \\
     &= \frac{1}{2}\iint \pygx ^2 \px \dx \dy 
     	+ \frac{1}{2}\iint \pygz ^2 \pz \dz \dy \\
     	&\ \ \ - \iint \pygz ^2 \pz \dz \dy.
\end{align*}
Let $\px = \pzzp$, and $\dx = \dz\dzp$. Then the final term can be expressed as
\begin{align*}
\iint  \pygz  ^2 \pz \dz \dy
   &= \iint \frac{\pzy}{\pz} \frac{\pzy}{\pz} \pz \dz \dy \\
   &= \iint \frac{\pzy}{\pz}\pzy \dz \dy \\ 
   &= \iint \frac{\pzy}{\pz} p(\bzp|\bz,\by) \pzy \dz \dzp \dy \\ 
   &= \iint \frac{\pzy}{\pz} p(\bz,\bzp,\by) \dz \dzp \dy \\ 
   &= \iint \frac{\pzy}{\pz} \pxy \dx \dy \\ 
   &= \iint \frac{\pzy}{\pz} \frac{\pxy}{\px} \px \dx \dy \\
   &= \iint \pygz \pygx \px \dx \dy,
\end{align*}
where $p(\bz,\bzp,\by) = \pxy$, and $\dz\dzp = \dx$ are used.
Therefore,
\begin{align*}
\widetilde{\mathrm{SCE}}(\bY|\bZ) - \widetilde{\mathrm{SCE}}(Y|X) 
     &= \frac{1}{2}\iint  \pygx  ^2 \px \dx \dy 
     	+ \frac{1}{2}\iint  \pygz  ^2 \pz \dz \dy \\
     &\ \ \ - \iint \pygz \pygx \px \dx \dy \\
     &= \frac{1}{2}\iint \left( \pygx - \pygz \right)^2 \px \dx \dy    	
\end{align*}
We can also express $\pygx$ in term of $\pygz$ as
\begin{align*}
\pygx &= \frac{\pxy}{\px} \\
      &= \frac{\pxy}{\px} \frac{\pzy}{\pzy} \\
      &= \frac{\pxy \pzy}{ \pzpgz \pz \pygz \pz } \\
      &= \frac{\pzzpy \pzy}{\pzpgz \pz \pygz \pz} \\
      &= \frac{p(\bzp,\by|\bz) \pzy}{\pzpgz \pygz \pz} \\
      &= \frac{p(\bzp,\by|\bz) }{\pzpgz \pygz } \pygz
\end{align*}
Finally, we obtain 
\begin{align*}
\widetilde{\mathrm{SCE}}(\bY|\bZ) - \widetilde{\mathrm{SCE}}(\bY|\bX) 
     &= \frac{1}{2}\iint \left( \pygx - \pygz \right)^2 \px \dx \dy    	\\
     &= \frac{1}{2}\iint \left( \frac{p(\bzp,\by|\bz) }{\pzpgz \pygz } \pygz - \pygz \right)^2 \px \dx \dy    	\\ 
     &= \frac{1}{2}\iint \left( \frac{p(\bzp,\by|\bz) }{\pzpgz \pygz } - 1 \right)^2 \pygz^2 \px \dx \dy    	\\     
     &\geq 0,
\end{align*}
which concludes the proof.

\section{Derivatives of SCE}
\label{appendix:derivative-SCE}
Here we show the formula of derivatives of $\hSCE(Y|Z)$ using LSCE estimator.
SCE approximation by LSCE estimator is 
\begin{align}
\hSCE(\bY|\bZ)  
			&= \frac{1}{2} \bhalpha^\top \bhG \bhalpha - \bhh^\top \bhalpha. \nonumber
\end{align}
Taking its partial derivatives with respect to $\bW$ and we obtain
\begin{align}
\pSCEW 
			&= -\frac{1}{2} \frac{\pa \bhalpha^\top \bhG \bhalpha}{\paW} - \frac{\pa \bhh^\top \bhalpha } {\paW} \notag \\
			&= \frac{1}{2}\left( \frac{\pa \bhalpha^\top}{\paW}\bhG\bhalpha + \frac{(\bhG\bhalpha)^\top}{\paW}\bhalpha\right)	- \frac{\pa \bhalpha^\top} {\paW} \bhh - \frac{ \pa \bhh^\top}{\paW} \bhalpha \notag \\
			&= \frac{1}{2}\frac{\pa \bhalpha^\top}{\paW}\bhG\bhalpha + \frac{1}{2}\frac{\pa \bhalpha^\top}{\paW}\bhG\bhalpha + \frac{1}{2} \bhalpha^\top\frac{\pa \bhG}{\paW}\bhalpha 
			- \frac{\pa \bhalpha^\top} {\paW} \bhh - \frac{ \pa \bhh^\top}{\paW} \bhalpha \notag \\
			&= \frac{\pa \bhalpha^\top}{\paW}\bhG\bhalpha 
			+ \frac{1}{2} \bhalpha^\top\frac{\pa \bhG}{\paW}\bhalpha 
			- \frac{\pa \bhalpha^\top} {\paW} \bhh - \frac{ \pa \bhh^\top}{\paW} \bhalpha.
			\label{Append:SCE_deriv}
\end{align}
Next we consider the partial derivatives of $\bhalpha$ as follows
\begin{align}
\frac{\pa \bhalpha}{\paW} &= \frac{\pa (\bhG + \lambda \bI)^{-1} \bhh}{\paW} \notag \\
         &= \frac{\pa (\bhG + \lambda \bI)^{-1}}{\paW} \bhh + (\bhG + \lambda \bI)^{-1} \frac{\pa \bhh}{\paW} \notag \\
\frac{\pa \bhalpha^\top}{\paW}  &= (\frac{\pa (\bhG + \lambda \bI)^{-1}}{\paW} \bhh)^\top + \frac{\pa \bhh^\top}{\paW} (\bhG + \lambda \bI)^{-1}. 
\label{Append:SCE_alpha_deriv}
\end{align}
Using $\frac{\pa \boldsymbol{X}^{-1}}{\pa t} = -\boldsymbol{X}^{-1} \frac{\pa \boldsymbol{X}}{\pa t} \boldsymbol{X}^{-1}$, we obtain
\begin{align}
\frac{\pa (\bhG + \lambda \bI)^{-1}}{\paW} \bhh   
   &= -(\bhG + \lambda \bI)^{-1} \frac{\pa \bhG}{\paW} (\bhG + \lambda \bI)^{-1} \bhh - (\bhG + \lambda \bI)^{-1} \frac{\pa \lambda \bI}{\paW} (\bhG + \lambda \bI)^{-1} \bhh \notag \\
   &= -(\bhG + \lambda \bI)^{-1} \frac{\pa \bhG}{\paW} \bhalpha  - 0 \notag \\
(\frac{\pa (\bhG + \lambda \bI)^{-1}}{\paW} \bhh )^\top  &= -\bhalpha^\top \frac{\pa \bhG}{\paW} (\bhG + \lambda \bI)^{-1}. \label{Append:SCE_Qi_deriv}
\end{align}
Substitute Eq.\eqref{Append:SCE_Qi_deriv} into Eq.\eqref{Append:SCE_alpha_deriv} to obtain
\begin{align}
\frac{\pa \bhalpha^\top}{\paW} 
   &= -\bhalpha^\top \frac{\pa \bhG}{\paW} (\bhG + \lambda \bI)^{-1} + \frac{\pa \bhh^\top}{\paW} (\bhG + \lambda \bI)^{-1}. \label{Append:SCE_alpha_Qi_deriv}
\end{align}
Finally, by substitute Eq.\eqref{Append:SCE_alpha_Qi_deriv} into Eq.\eqref{Append:SCE_deriv} and use $(\bhG + \lambda \bI)^{-1} \bhG \bhalpha = \bhbeta$, we have
\begin{align*}
\pSCEW  	&= -\bhalpha^\top \frac{\pa \bhG}{\paW}\bhbeta 
			+ \frac{\pa \bhh^\top}{\paW} \bhbeta + \frac{1}{2}\bhalpha^\top \frac{\pa \bhG}{\paW}\bhalpha 
			\notag \\
  & \ \ \ \ + \bhalpha^\top \frac{\pa \bhG}{\paW}\bhalpha 
  		    - \frac{\pa \bhh^\top}{\paW} \bhalpha - \frac{\pa \bhh^\top}{\paW} \bhalpha
			\notag \\
			&= \bhalpha^\top \frac{\pa \bhG}{\paW} (\frac{3}{2}\bhalpha - \bhbeta)
			+ \frac{\pa \bhh^\top}{\paW} (\bhbeta - 2\bhalpha),
\end{align*}
where the partial derivatives of $\bhG$ and $\bhh$ depend on the choice of basis function. 

Here we consider the Gaussian basis function described in Section \ref{section:model}.
Their partial derivatives are given by
\begin{align*}
\frac{\pa \hQ_{k,k'}}{\paW}
    &= -\frac{1}{\sigma^2n} \sum_{i=1}^n\bar{{\Phi}}_{k,k'}(\bz_i) 
     \left( (\bz^{(l)}_i - \bu^{(l)}_k) (\bx^{(l')}_i - \but^{(l')}_k) + (\bz^{(l)}_i - \bu^{(l)}_{k'}) (\bx^{(l')}_i - \but^{(l')}_{k'}) \right)  \\
\frac{\pa \hq_k}{\paW} 
	&= -\frac{1}{\sigma^2n}\sum_{i=1}^n \varphi_k(\bz_i, \by_i) \left( (\bz^{(l)}_i - \bu^{(l)}_k) (\bx^{(l')}_i-\but^{(l')}_k) \right).
\end{align*}
\end{document}